Dafydd Gibbon and Sascha Griffiths

# Multilinear Grammar: Ranks and Interpretations

**Abstract:** Multilinear Grammar provides a framework for integrating the many different syntagmatic structures of language into a coherent semiotically based Rank Interpretation Architecture, with default linear grammars at each rank. The architecture defines a *Sui Generis Condition* on ranks, from discourse through utterance and phrasal structures to the word, with its sub-ranks of morphology and phonology. Each rank has unique structures and its own semantic-pragmatic and prosodic-phonetic interpretation models. Default computational models for each rank are proposed, based on a *Procedural Plausibility Condition*: incremental processing in linear time with finite working memory. We suggest that the Rank Interpretation Architecture and its multilinear properties provide systematic design features of human languages, contrasting with unordered lists of key properties or single structural properties at one rank, such as recursion, which have previously been been put forward as language design features. The framework provides a realistic background for the gradual development of complexity in the phylogeny and ontogeny of language, and clarifies a range of challenges for the evaluation of realistic linguistic theories and applications. The empirical objective of the paper is to demonstrate unique multilinear properties at each rank and thereby motivate the Multilinear Grammar and Rank Interpretation Architecture framework as a coherent approach to capturing the complexity of human languages in the simplest possible way.

## 1 Introduction: linearity, ranks, interpretations[1]

### 1.1 Procedures and ranks

Frameworks for language description in linguistics and the human language technologies tend to concentrate on single components, such as a syntactic model, a morphological or phonological model, models for narration or argumentation in texts, or models of discourse sequencing. Each model tends to have a different formal basis. The result of this specialisation is that language modelling as a whole turns out to be a hybrid system with interfaces between component subsystems. It is not impossible that this concept provides an adequate approach to understanding human language abilities and to language development in acquisition or evolution, subsystem by subsystem. But in terms of Occam's Razor, the economy of descriptive and explanatory means, it is a wasteful strategy. There are recent – or revived – developments in language description which are based on formally integrated and semiotically motivated approaches. In order to combine formal integration and semiotic motivation we propose a rank hierarchy of domains from discourse to word, composed of default linear structures and their semantic-pragmatic and prosodic-phonetic interpretations at each rank. The overriding aim is to formulate a foundation for *combinatorial communication* (Scott-Phillips and Blythe 2013).

In the first development, the structures of language, particularly of speech, spoken language, are increasingly being grounded in an empirically motivated procedural turn (Christensen and Chater 2016; List et al. 2016). This development puts new life into the traditional linguistic perspective of the epistemological primacy of gestural and speech data over written data. The notion of procedure in this context means, essentially, an algorithm or computation which operates in time and memory space, and is not to be confused with performance as used in the Chomskyan sense to refer to any vagaries of actual language use. Procedures

1 This study is dedicated to the memory of Hans-Jürgen Eikmeyer, colleague and teacher, to whose innovative ideas and profound knowledge of computational methodologies both authors are heavily indebted. We are very grateful to our anonymous reviewers for their patient and careful reading of the first version and for their numerous constructive comments, which have helped to clarify the structure and argumentation of our study, and to Damir Cavar, Jue Yu, Laura Liu and the Bielefeld linguistics colloquium audience for comments on earlier versions and presentations.

are just as abstract, selective and systematic as generative or constraint-based competence theories, and the present approach can be thought of as carving out one systematic aspect of the fuzzy performance domain and re-integrating concepts of gestured, spoken and written communication into theories and models of language. The heterogeneous varieties of language require a principled and systematic treatment of procedurality as a plausibility constraint on realistic grammars. The relevant procedural properties are linearity and incrementality, based on grammars which can be processed in linear time with finite working memory. It seems odd, from a procedural point of view, that the 'langue' (de Saussure), 'Sprachgebilde' (Prague School) and 'competence' (Chomsky) linguistic frameworks use static category, feature and rule combinations which abstract away from criteria of time and movement, except in lexical and phrasal semantics, and in duration features and stress positions in phonology.

Second, attention is continuing to turn to the broader semiotic context, beyond the traditional linguistic focus on the word, its internal structure of morphology and phonology, its external distribution in the sentence and its prosodic-phonetic and semantic-pragmatic interpretations. In recent terminology, the 'external language' of prosodic-phonetic interpretation, and the cultural context of semantic-pragmatic interpretation enter into a semiotic relation mediated by linear structures. The ranks of discourse and text, along with their distinct processing requirements and semiotic properties of prosodic-phonetic and semantic-pragmatic interpretation, have in general been seen in linguistics and the language technologies as being quite separate from the ranks of sentence and word. Likewise, until relatively recently the gestural communication modality has also been seen to be quite separate from speech and writing, disregarding the fact that speech and writing are also grounded in vocal and manual gestures, respectively. But the principled incorporation of the 'larger' domains of language structure is coming more into focus, with properties such as longer time windows (Christensen and Chater 2016), in prosodic-phonetic interpretation (Tillmann and Mansell 1980; Gibbon 1992; Carson-Berndsen 1998) as well as in semantic-pragmatic interpretation of the higher ranks in cultural contexts and in multimodal discourse (Gibbon 2011).

We explore properties of each domain in terms of a semiotically motivated multidimensional Rank Interpretation Architecture, in which each rank from discourse through utterance and text to phrase and word has a ternary configuration with its own unique categories, functions and structures and its own parallel semiotic pair of prosodic-phonetic and semantic-pragmatic interpretations. Prosody, for example, covers the phonetic interpretations of ranks above constituents of monosyllabic words, idioms are lexical items at higher ranks than the word, and narration and argument constitute semantic-pragmatic interpretations at the higher ranks.

One very simple informal example of a rank triple <*structure, prosodic-phonetic interpretation, semantic-pragmatic interpretation*> at the discourse level is an unanswered call with chanted intonation (details to be specified at lower ranks):

1. structure: [ CHANT: call, RESPONSE: _ ] (understroke indicates an uninstantiated variable);
2. prosodic-phonetic interpretation: [ PHON: *wɛər¯aː–juː* ] (horizontal bars indicate level pitch heights; an orthographic interpretation [ ORTH: *Where are you?* ] or a gestural interpretation (such as waving) can be added, generalising prosodic-phonetic interpretation to a multimodal surface interpretation [ SURF: [ GEST: …, PHON: …, ORTH: … ] ]);
3. semantic-pragmatic interpretation: a situation frame with speaker, uninstantiated addressee variable, and a dialogue act context for initiating communication (see also section 4.3).

This network of ranks and interpretations with default linear grammars at each rank characterises human speech and its heterogeneity more realistically than unstructured lists of design features (Hockett 1958) or selected formal properties at a single rank such as phrase structure recursion (Hauser et al. 2002). Realistic approaches depend, among other criteria, on a *Procedural Plausibility Condition* of linear processing time and finite working memory. Some grammar types with unrestricted memory and nonlinear processing time, such as context-free or phrase structure grammars, do not share these properties. Constructions which involve centre-embedding can be characterised as generalisations from a linear right-branching (or left-branching) context to a central context. The resulting more complex conditions on processing are made possible, at least to a limited extent, by enhanced but still restricted time and memory resources such as those available in writing and rehearsed speech. From the perspective of acquisition (ontogeny), evolution (phylogeny) or simply language history, centre-embedded constructions thus represent a bold step, for example from right branching rhematic object relative clauses in an SVO language to centre-embedded thematic relative clauses in subject position, forcing a construction for which human speech processing facilities are inadequate. Even so, for centre-embedding the finite working memory condition still applies, a condition fulfilled by the augmented linear grammars of Langendoen (1975) and corresponding to the memory limitation requirements noted by Kindt (1998) and Karlsson (2010).

The Rank Interpretation Architecture is multidimensional. Multidimensionality applies first to the hierarchy of *ranks*, second to the semiotic pair of prosodic-phonetic and semantic-pragmatic *interpretations* at each rank, and third to the parallel processing of simultaneous *streams* of input and output for these ranks and interpretations. Related multidimensional approaches, generally with more limited domains, are found in Firthian polysystemic linguistics (Firth 1948, Firth 1957; Palmer 1968, Palmer 1969), in Pike's tagmemics (1967), in Lamb's stratificational grammar (1966), in autosegmental phonology (Goldsmith 1976, Goldsmith 1990) and in articulatory phonology (Browman and Goldstein 1989), as well as in many studies of conversational gesture (cf. McNeill 2000; Kendon 2004; Rossini 2012). At discourse and utterance ranks the multidimensionality principle applies to the co-processing of co-expressive prosody-locution relations, with prosody as a metalocutionary channel of information denoting positions and properties of lexico-syntactic locutions (Gibbon 1983). From an empirical point of view, each rank must be justified by a *Sui Generis Condition*, that is, by demonstrating that the rank has unique properties which do not apply in the same way to other ranks. Inevitably there are (over)simplifications in a study of this kind, but the main idea is to argue for an overall coherent framework rather than for a hybrid framework with *ad hoc* interfaces.

In the following parts of the present section the Multilinear Grammar and Rank Interpretation Architecture framework is outlined. Sections 2 and 3 describe Multilinear Grammar and the Rank Interpretation Architecture, respectively, in more detail. Sections 4, 5, 6 and 7 discuss empirical aspects of the discourse, utterance, phrase and word ranks, and Section 8 formulates a summary, conclusions and the outlook for further developments.

## 1.2  Flat grammars

Over the past few decades there have been occasional moves towards 'flat grammars' based on the plausibility criteria of linear processing time and finite working memory, particularly in connection with the readjustment of locutionary structures to the linearity of prosody, and in computational morphology and phonology. In these approaches, to be discussed in the later sections of this study, default patterns at each rank are captured by left branching or right branching grammars or equivalently by finite state automata. Readjustment and restructuring rules were introduced in order to achieve linearisation and empirically more realistic structural depth-reduction of complex grammatical structures (Bierwisch 1966; Langendoen 1975; Culicover and Rochemont 1983; Nespor and Vogel 1983). If structures are recognised as linear in the first place, readjustment and restructuring rules are in principle unnecessary. The flat grammar paradigm proposes procedural plausibility criteria for syntax and for prosodic-phonetic interpretation, but not necessarily for semantic-pragmatic properties, which need to describe complex cognitive processes and cultural interactions, from music and dance to town-planning or computer chip design. An explicit linguistic approach to the linearisation of semantic structures was described by Chafe (1970).

The flat grammar approach is justified empirically, but also arises from applying Occam's Razor, the simplicity principle, in two ways. First, powerful centre-embedding recursion and cross-serial formalisms such as context-free and indexed grammars have unbounded working memory requirements, which is an empirically implausible assumption in view of the limited working memory and linear real-time processing of speech, and even of the more relaxed constraints on writing. Overly powerful formalisms are being abandoned in many contexts, either in favour of flat linear grammars, or in favour of more complex grammars which are nevertheless constrained to have only finite working memory requirements. These issues are discussed in more detail in Section 2. The many grammatical constraints which only affect relations within 'root small clauses' (Progovac 2010), such as argument relations and argument ordering, are inherently linear and fall within the domain of finite structures and finite working memory. Apparent exceptions, such as *wh*-fronting, are discussed in Section 6.

Second, it remains uncontested that semantic interpretation and disambiguation may sometimes require arbitrarily complex structures in order to express the full complexity of general cognition and cultural context. But this does not necessarily apply to grammar in the narrow sense, which basically needs to account for the combinatorial facts of language varieties. A thought experiment may help to clarify. Imagine someone attending a lecture in a subject in which they have no training: the grammar is recognisable, but it is largely devoid of content for that listener, except for  structurally local contributions of inflectional morphology, articles, auxiliary verbs, prepositions and conjunctions.

The properties of flat grammars, as expressed in regular grammars and finite state automata, have been well investigated in terms of their algebraic and parsing and generation properties (Langendoen 1975; Kaplan and Kay 1994) as well as their learnability (Berwick and Pilato 1987). The right branching and left branching grammars which figure prominently in linguistic descriptions can be implemented without loss by finite state automata (Hopcroft et al. 2007). The relevant procedural properties of algorithms for processing these, and more complex, grammars are detailed by Kay (1980). In computational contexts these results are well-known, but in linguistic discussions they tend to be sidelined by Chomsky's early theorem (1957:21, example 9): *English is not a finite-state language*. Chomsky's theorem applies – with certain restrictions – to the complexities of written language and rehearsed speech, because these varieties have at their disposal additional memory resources and more leisurely time windows which to some extent support the additional temporal and storage requirements of centre-embedded constructions. But Chomsky's theorem does not apply to non-practised spontaneous speech which is produced and understood on the spur of the moment, is restricted to a working memory with a rather small finite bound (Church 1980) and is error-prone on more complex tasks.

In recent years, the status of flat grammars has figured in discussions of types of recursion in phrasal syntax, partly in the context of the controversial claim (Hauser et al. 2002; Fitch et al. 2006) that recursive hierarchy formation ('merge') is the defining property of human languages. This claim has been extensively discussed in contributions to the van der Hulst (2010) and Roeper and Speas (2015) collections. In the Multilinear Grammar approach, incremental linear syntactic composition and highly restricted forms of left branching or right branching recursion, not general centre-embedding recursion, are taken to be the default cases in human languages, particularly in gestural and speech communication. In marked, non-default cases with the limited centre-embedding recursion generalisation, it is still the case that only very limited finite working memory is available (Langendoen 1975; Kindt 1998; Karlsson 2010). A principled description of the heterogeneity of language varieties on the basis of both structural and procedural plausibility criteria is required for realistic grammars, and there is no reason to maintain a distinction between 'internal' or competence grammars with unlimited working memory and 'external' or performance grammars with limited working memory if the goal is empirical realism. This is not to say that it makes sense to postulate a simple correspondence between human behaviour in time and the procedural properties of grammars and their processing algorithms in time, as in older psycholinguistic concepts of realistic grammars. Consideration of formal procedural criteria, from basic search algorithms through stochastic optimisations to contemporary deep neural learning algorithms, has nevertheless long been considered to be a necessary first step in the development of realistic grammars (Levelt 1974).

Hierarchical, cross-referencing and other *prima facie* nonlinear syntagmatic structures are relevant for semantic compositionality, anaphora and information structure and the context of speaking, but it can be shown that this is not necessarily the case for the grammar itself. These nonlinear structure types are more complex than parallel flat linear patterns and need more complex knowledge access mechanisms. Even when they have been postulated for grammars, they have been ultimately motivated by pragmatic, semantic and general cognitive representation requirements, and need to be reconstructed from flatter grammatical structures with the aid of context. In the description of grammar itself, the more complex structures are convenient conceptual tools for representing generalisations about similar subpaths along the flat paths defined by linear grammars and finite state machines. These generalisations are motivated by aspects of semantic-pragmatic interpretation and not *per se* proof of the need for more complex grammar types. Clearly, production and perception at all ranks in both speech and writing are also limited by finite naturally or conventionally determined quantities of time and energy, from phonetics to Gricean maxims. But this is a separate issue.

The default procedural plausibility criteria of linear time and finite working memory are at the very least a useful starting point and a potentially falsifiable heuristic benchmark, perhaps even a necessary theoretical core. As complexity increases during the ontogeny and phylogeny of gestural, spoken and written communication, increasingly complex linear patterns appear, from the simplest sound sequences through the basic clausal patterns proposed by Wittenberg and Jackendoff (2014) and Jackendoff and Wittenberg (2016) to the centre-embedding generalisation with its processing problems. We claim that the centre-embedding emerges as a generalisation of non-centre-embedding recursion on the different time scales of spontaneous speech improvisation, of language acquisition, of the history of language change and of language evolution, a claim which is speculative but not implausible, and far simpler and potentially easier to falsify than claims about the triggering of recursion by genetic mutation (though not excluding this possibility).

Intonation patterning, like morphology and phonology at the word rank, can be also modelled in conformance with the *Procedural Plausibility Condition*. This has been known for a long time (Johnson 1972; Pierrehumbert 1980; Hirose et. al. 1984; 't Hart and Cohen 1990; Gibbon 1981a, 1987; Koskenniemi 1983; Kaplan and Kay 1994). Other studies (Chomsky 1965; Reich 1969; Langendoen 1975; Pierrehumbert 1980; Culicover and Rochemont 1983; Nespor and Vogel 1983; Ladd 1996; Féry 2017) have confirmed the need for flat linear intonation models, sometimes introducing readjustment rules for linearisation and structure reduction, sometimes relegating such requirements to a vaguely defined performance domain. In morphology (particularly in compound word formation), as in phrasal syntax, tree hierarchies are still necessary, but for semantic disambiguation and compositional semantic interpretation, not for strictly linear combinatorial properties (see Sections 4 to 7).

The upshot of these arguments is that if distributional, combinatorial syntactic properties are separated from strictly semantic constraints, then linearity is the appropriate default for structure and process in speech, and hierarchical structuring is just needed for rather rare instances of centre-embedded recursion in speech, which only occurs with severe depth constraints (Section 6). The *Procedural Plausibility Condition* which justifies default flat grammars is valid at all ranks:

1. at discourse rank, for example for adjacency pairs, question-answer-confirmation triples and their iterations;
2. at utterance rank, to describe narration and argument patterns;
3. at phrasal rank, modelled by left branching or right branching structures;
4. at word rank and the morphology and phonology sub-ranks, for the linear combinatorics of inflectional and derivational affixation, compounding, and phonotactics.

We surmise that in the default case the semantic-pragmatic interpretations of these linear sequences can also be handled in a linear, incremental manner (Hough et al. 2015; Fernando 2004), though in specific cases semantic-pragmatic interpretation is more complex.

## 1.3   Processing issues

In addition to the arguments for the empirical grounding of realistic grammars in the *Procedural Plausibility Condition* of linear processing time and finite working memory, a number of other formal and informal arguments for linearity can be made. One key argument is the traditional Saussurean postulate of the primacy of spoken language versus writing, to which the primacy of gesture versus spoken language may be added. Gestures in speech production, gesticulation and signing events are time functions which require fast and globally near-deterministic processing of many parallel information streams (Browman and Goldstein 1989; Christensen and Chater 2016). Writing is more leisurely. Speech and gesticulation are ontologically quite different from the inscribed traces which are characteristic of written languages, which involve slower manual production and perception processes than speech, and are supported by additional paper and screen working memory. Many descriptive and theoretical linguistic studies have been built on the written data type with its enhanced working memory potential, often with a hermeneutic methodology using introspected data types which are imagined rather than observed. These data types are valid for some purposes, but studies based on them are not easily generalisable to the structures and real time processes of speech.

Another argument for procedural plausibility derives from psycholinguistic studies of the lexicon. In older studies, decision-tree-like lexical access patterns define cohorts of lexical items which are disambiguated incrementally as input continues (Marslen-Wilson 1987). The cohort approach has been replaced by formal paradigms such as deep artificial neural network classifiers or other statistical models, but the later approaches still capture key insights of incrementality and linearity.

A more circumstantial argument for the *Procedural Plausibility Condition* is based on external evidence: the success on theoretical and operational grounds of stochastic flat grammars in speech technology and of statistical finite state approximations to more complex context-free grammars (Pereira and Wright 1997) in natural language processing applications such as machine translation. Controlled studies on the use of flat linear *n*-gram contexts in phoneme, morpheme and word segmentation extend the basic flat linear modelling strategy (Church 2007; Griffiths et al. 2015b, 2015c). In these areas of the human language technologies, complementary properties and similarities between artificial neural networks and stochastic regular grammars such as Hidden Markov Models, essentially statistically weighted finite state automata and transducers, are exploited (Jurafsky and Martin 2009:233ff., 266ff.; Huang et al. 2001:377ff., 457ff.).

Briefly, Multilinear Grammar in the Rank Interpretation Architecture is characterised in terms of a small finite tuple of *ranks*, ranging from discourse through utterance and phrase to word and the sub-ranks of morphemes and phonemes (using the term 'phoneme' cross-theoretically, to mean a class of smallest word-distinguishing segments of speech). In the default case which is relevant for spontaneous speech, the structures at each rank are processed incrementally (cf. also Gaspers et al. 2011), in parallel, with finite working memory, and rather quickly, in linear time: about 10 phonemes, 5 syllables, 1.5 words, 0.5 sentences per second, with each rank being assigned its own characteristic semiotic pair of incremental semantic-pragmatic interpretation and prosodic-phonetic interpretation.

The crucial point for our purposes is whether each rank fulfils a *Sui Generis Condition*, with its own unique linear structures and its own unique physical prosodic-phonetic interpretations, which in turn fulfil the *Procedural Plausibility Condition* of finite working memory and linear processing time. We discuss selected experimental and corpus analyses in Sections 4, 5, 6 and 7 of the present study, but focus more on illustrating a coherent overall approach than on detailed experimentation or corpus analysis.

# 2 Realistic grammar: linearity and hierarchy

## 2.1 Grammar theories and models

The multidimensional Rank Interpretation Architecture framework has distinct semantic-pragmatic and prosodic-phonetic interpretations at each rank and thus differs markedly from traditional architectures of language as a semiotic system. One common traditional approach places semantics and pragmatics at the 'top' and phonetics at the 'bottom' of a cascade of levels with syntax in between. Another approach which is almost as common has pragmatics as an all-inclusive domain subsuming semantics, which in turn subsumes syntax (which presumably includes morphology and phonology). In the Rank Interpretation Architecture, however, each rank in the hierarchy has its own ternary semiotic make-up of structures and interpretations. Each rank has not only its own flat default grammar but also its own lexicon as an inventory of signs, from phonemes and morphemes to conventional lexemes, and from phrasal idioms and stereotypic utterances to textual elements from proverbs to literary texts and from ritual components of discourse such as greetings and stereotypic small talk to religious liturgies and formal legal discourse.

There are many logics, algebras, geometrical calculi and other syntaxes for defining grammars. The most popular type is the conventional string rewriting grammar formalism which is frequently used in linguistics, whose formal properties are well understood. In order to position the arguments sufficiently precisely for both descriptive and computational interests, the most relevant properties of string rewriting grammars are outlined, focussing on the procedural plausibility criteria of processing time, defined as a linear function of the length of the input, and finite working memory. In general, linguistic rather than algebraic or logical terminology is used, without the usual formal definitions and proofs, which are accessible elsewhere (Hopcroft et al. 2007).

Following one common terminological convention, a grammar $G$ for a language $L$ is a *theory* consisting of

1. a set of *general premises* (the grammar rules),
2. a *specific premise*, i.e. a minimal string consisting of a single symbol, such as $A$.
3. an inference procedure which uses the specific premise as an *initial theorem*, to infer a set of *derived theorems* from the general premises by modus ponens until no more general premises apply, finally yielding sets of theorems such as {*big*, *small*, *very big*, *very small*, *very very big*, *very very small*, …}. The inference procedure is very familiar to linguists.

Continuing with the minimal example, a set of general premises which permit inferring this particular set of theorems from the initial specific theorem includes:

1. general theorems represented as string rewriting rules $A{\rightarrow}C$ and $A{\rightarrow}B \frown A$, with the '$\rightarrow$' rewrite operator and the tie '$\frown$' string concatenation operator, and with $A$ as the initial specific premise;
2. general theorems represented as lexical rules $B$ = {*very*}, $C$ = {*big*, *small*}, and the '=' operator, with the set union $B \cup C$ of lexical sets representing the lexicon.

Inference proceeds by replacing any occurrence of the left-hand side of a rule in a theorem with the right-hand side of that theorem (for the '$\rightarrow$' operator), or an element of the set on the right-hand side of the theorem (for the '=' operator). Ignoring a few details, terminal theorems are inferred by selecting matching resulting theorems with the general premises until no more substitutions are possible and all elements in the resulting theorem are elements of the lexicon.

A *theory* such as the grammar $G$ is understood here as a description of a *model*: an interpretation function maps theorems of the grammar to a domain of structures, either abstract structures or structures representing observed reality. The *model* is thus an interpretation of the theory. Prosodic-phonetic interpretations in terms of externally observable physical events at each rank are one kind of model for the part of the theory which pertains to that rank. The semantic-pragmatic interpretations are a second kind of model. A third kind of model for the theory is a set of rooted ordered tree graphs: the phrase structures of linguistic analysis, reflecting the theorem inference process, with the initial theorem and the elements of the right-hand sides of the '→' premises as nonterminal nodes, and leaf nodes corresponding to elements of the sets in the '=' premises.

A fourth type of model for a theory is a *procedural model*, expressed in terms of specific algorithms and data structures, and possibly implemented as a computer program. Depending on the type of premise, different types of procedural model may be defined. For instance, a purely right branching or purely left branching grammar (but not a mixture of the two types) can be mapped to a finite state automaton model and straightforwardly implemented in a programming language.

Without going into details, the elementary example grammar given above can be mapped into the equally elementary finite state automaton model represented in Table 1. Each move of the automaton is a transition along a path from a source symbol (or node) in the leftmost column through an input or output symbol in a cell on its right to a target symbol (or node) heading the column which contains the selected cell. The target symbol then becomes the new source symbol, unless it is the final symbol. As an automaton moves along a transition it can process (emit or accept) the string it encounters along the way; the automaton is neutral with regard to production or reception.

Table 1: Finite state automaton in table notation as procedural model for grammar G.

|   | A | END | END |
|---|---|---|---|
| A | *very* | *big* | *small* |

The automaton shown in Table 1 can move along three transitions: starting with state *A* (in the leftmost column), the automaton may take a transition to the *END* state at the top of the rightmost column, and process either the word *big* or the word *small*, taken from one of the cells shared by the row with *A* and the columns with *END*. If the third option of looping back to itself is taken (to *A* at the top of the middle column), the automaton will process the word *very*. The <A, *very*, A> iteration (in tuple notation) can stop at any time and take the transition <A, *big*, *END*> or <*A*, *small*, *END*> to the final state. In this way, iterated sequences of arbitrary length can be accepted or emitted.

There are several notations for expressing automaton models in addition to the tabular and tuple notations. For example the regular expression *very\* big | small* describes the same language. Transition diagrams are often used (Figure 6 shows a more complex example). The formalisms are equivalent in that a finite automaton describing a specific language, i.e. a set of strings, may be represented by any of these notations.

Accepting or emitting an unlimited number of strings is overkill for a realistic grammar, of course, and in any automaton implementation the number of loop transitions which can be traversed is limited by the amount of available time and energy (like the exhalation and articulation rates and durations of a human speaker).

## 2.2   Regular grammars and their properties

The conventions used in the semi-formal definition of grammar *G* in the preceding subsection are taken from the Chomsky-Schützenberger hierarchy of formal languages and formal grammars (Chomsky and Schützenberger 1963; Hopcroft et al. 2007). The grammar *G* represents the most restrictive type of grammar in the hierarchy, the *Type 3 grammar* or *regular grammar*, which has either only right branching rules or only left branching rules but not a mixture of both. The language *L* is a *regular language*. Omitting the explicit concatenation operator '⌢', rules in a regular grammar have the form of terminal or lexical rules $A{\rightarrow}a$ and concatenation rules, either of the right branching format $A{\rightarrow}aB$ or the left branching format $A{\rightarrow}Ba$. *A* and *B* are nonterminal symbols and occur in the left-hand sides of rules, and *a* is a terminal symbol. Any right branching regular grammar $G_{RB}$ can be converted into a left branching regular grammar $G_{LB}$ modelled by the same finite state automaton and describing the same language: $L_{GRB} = L_{GLB}$.

The finite state automaton model can be generalised as a finite state transducer, which processes tuples of symbols (usually pairs), rather than single symbols, and thereby describes regular relations, which can be thought of as translations between regular languages. Finite state transducers are commonly used for prosodic-phonetic interpretation in computational morphology and phonology.

Two main types of regular grammar can be distinguished in developing a realistic linguistic grammar. It will be useful to be able to refer to these in the later discussion:

1. *acyclic* regular grammars, which (given a finite vocabulary) describe a finite set of strings $L_{acyclic}$, and have acyclic finite state automata as procedural models;

2. *cyclic* or *recursive* regular grammars (like grammar $G$ in the example), which are either right branching (*tail recursion*) or left branching (*head recursion*), describe an infinite set of strings $L_{cyclic}$, and have cyclic or *iterative* finite state automata as procedural models.

Cyclic regular grammars are recursive: a given rule may apply to its own output, either if its left-hand symbol occurs on its own right-hand side (direct recursion) or on the right-hand side of a string created by substitution later in the derivation (indirect recursion). Right branching recursion is also termed *tail recursion*, and left branching recursion is termed *head recursion* (though head recursion is sometimes confusingly referred to as tail recursion in the linguistic literature, perhaps because 'head' has other meanings such as the category-determining element of an endocentric construction). Recursion in a finite state automaton, i.e. iteration, is expressed as loops from a state back to itself (direct iteration) or to a previous state (indirect iteration).

There are several criteria for choosing between head recursive and tail recursive grammars. One criterion involves semantic interpretation: a right branching regular grammar may reflect the desired compositional semantic modification hierarchy, and possibly also the reflexes of information structure in prosody, more closely than a left branching regular grammar, or vice versa. A significant procedural criterion is that top-down and bottom up processing schedules have different advantages and disadvantages with respect to left and right branching grammars (Jurafsky and Martin 2009 Chapter 13). A finite state automaton model has no branching in the same sense, so this problem does not arise, but the same distinctions can be expressed in terms of incremental processing conventions. There are other more technical properties of finite state automata which do not warrant treatment here.

One well-known case of a structure which can be described with an acyclic grammar or finite state automaton is the *strictly layered hierarchy*, that is, a non-recursive hierarchy, with well-defined left-hand side and right-hand side category types, which has finite depth, such as the prosodic hierarchy of Selkirk (1984) and its later variants. The rank hierarchy in the Rank Interpretation Architecture is also a strictly layered hierarchy, related to though not identical with the prosodic hierarchy. Although the rank hierarchy has finite depth, the grammars at a particular rank may be cyclic, as in the prosodic hierarchy, giving the entire system cyclic properties. Another example of a finite depth hierarchy (though not usually strictly layered in the same sense) is found in 'kernel' or 'simple active affirmative declarative' sentences or 'root small clauses' containing no recursion: the maximum length of the strings concerned is finite, the language consisting of the set of these sentences is finite, the grammar is acyclic, the tree graphs modelling their structure have finite depth, and the finite state automaton model has no loops. Another example is the set of syllables of a language: syllables have finite maximum length, finite depth and can be described by an acyclic regular grammar or a loop-free finite state automaton.

## 2.3 More complex grammars

The next less restricted languages in the Chomsky-Schützenberger formal language hierarchy after the regular languages are the *context-free languages*, or Type 2 languages, described by context-free (phrase structure, constituent structure, Type 2) grammars. Context-free languages have structural models in the form of rooted tree graphs which can branch in any direction, and procedural models in the form of *push-down automata*, which in principle require unrestricted working memory and nonlinear processing time. The main linguistically relevant property of context-free languages is that they capture unrestricted centre-embedding recursion, illustrated by arbitrarily complex sentences of the following type: *the man who the boy who the dog bit called fetched the police*. Centre-embedding beyond a depth of two is not too easy to process under the real-time conditions of speech, as it relies on unlimited working memory to link left and right contexts of arbitrarily deep centre-embeddings: *the man...fetched the police*, *who the boy...called*, *who the dog bit*. This problem does not occur with unidirectionally branching grammars. If a centre-embedded sentence like *the zebra whose skin, which a man from Orlando bought illegally, got lost was the last of its species* is analysable, it is because the semantic-pragmatic interpretations with their cultural associations at each level of embedding are very different, and because writing provides additional working memory on the page, and additional processing time. It is not not because the syntax itself has intrinsically tractable combinatorial properties. A sentence like *the man whose car the man who the other man saw saw saw the man* is puzzling, to say the least; there is no support for centre-embedding from phrasal syntax alone.

There are paraphrasing strategies for avoiding centre-embedding in the flow of speech. Passivisation, which reverses the order of constituents, can have this effect: *the police were fetched by the man who was called by the boy who was bitten by the dog*. This version can be described with a right branching regular grammar or a finite state automaton and is easy to process. Passivisation can even rescue the puzzling sentence *the man whose car the man who the other man saw saw saw the man*, at least to some extent: *the man was seen by the man whose car was seen by the man who was seen by the other man*. Dislocation or extraposition constructions can also be construed as centre-embedding avoidance strategies, with a similar effect of reducing working memory load by introducing unidirectional recursion instead of centre-embedding. Traditionally the strategies which locate longer items at the end of a sentence in this way are described by Behaghel's *Law of Increasing Constituents*, *das Behaghelsche Gesetz der wachsenden Glieder* (Behaghel 1909).

Despite their lack of procedural plausibility, context-free grammars are popular in linguistics because of their versatility in providing arbitrary branching. Nevertheless, very many linguistic grammars are in practice restricted to right-branching or left-branching constructions, which therefore fulfil the *Procedural Plausibility Condition*. Among the centre-embedding languages, the most popular kind is the binary branching type known in formal grammar theory as *Chomsky Normal Form*, which has only rules of the format $A{\rightarrow}a$ and $A{\rightarrow}BC$, where $a$ is a lexical item ('terminal symbol') and $A$, $B$ and $C$ are category symbols ('non-terminal symbols'). The regular languages are a well-defined subset of the context-free languages, and the regular grammars and finite state automata are special cases of the more general context-free grammars and push-down automata, respectively.

Context-free languages, and the grammars and automata which describe them, as already noted, have nonlinear temporal processing complexity: worst case processing time with the most efficient algorithms may have approximately cubic complexity, that is, processing time varies with approximately the cube of the length of the input, $n^3$. Simpler backtracking algorithms may even require exponential time such as $2^n$ for nondeterministic grammars, which contain rules which share the same left-hand side symbol and also have the first symbol on their right-hand sides. This results in a processing ambiguity. But nondeterministic regular grammars are reducible to deterministic regular grammars, thus rescuing the procedural plausibility criterion. The same applies to nondeterministic finite automata, but not to general context-free grammars or their nondeterministic pushdown automata models. Spatial complexity for processing context-free grammars is such that in principle they require unrestricted working memory as well as nonlinear processing time, and thus fail the procedural plausibility criterion.

If centre-embedding is intrinsically hard to process, what could be its origin in the restricted centre-embeddings found in speech? Three possible assumptions about the genesis of centre-embedding are:

1. Centre-embedding in speech and writing has its origin in the structural generalisation of right (or left) branching constructions (such as an object relative clause with rhematic or comment

2. The centre-embedding generalisation is enabled by the registers of writing and rehearsed speech, which have enhanced paper or screen working memory capacity and allow increased processing time.

3. Failures of centre-embedding in speech are expected, because of lack of adequate working memory and processing time.

Development from unidirectional embedding based on regular grammars to more complex centre-embedding based on context-free grammars is not entirely implausible. The suggestion fits well with suggestions made by Wittenberg and Jackendoff (Wittenberg and Jackendoff 2014; Jackendoff and Wittenberg 2016) on the complexification of simpler clauses during the phylogeny and ontogeny of language.

The Chomsky-Schützenberger hierarchy contains grammar types which are more complex than Type 2 context-free grammars, the next more complex stage being Type 1, the context-sensitive grammars. In context-sensitive grammars, the general '→' premises have the form $XAY{\rightarrow}XBCY$: both the left-hand side and the right-hand side of the expression $A{\rightarrow}BC$ have an explicit left context $X$ and a right context $Y$, unlike the context-free rules. Context-sensitive rules like these are often abbreviated in linguistic descriptions as $A{\rightarrow}BC/X\_\_Y$, especially in phonology. Phonological 'context-sensitive' rules are typically not context-free in this technical sense but have procedural models which define the phonology-phonetics mapping by means of finite state transducers with input-output symbol pairs rather than with single input or output symbols (Johnson 1972; Koskenniemi 1983; Kaplan and Kay 1994; Beesley and Karttunen 2003).

Context-sensitive grammars are typically used at the phrasal rank in order to account for cases which require more complex recursive constructions than tree graphs can represent, such as *cross-serial dependencies* in sentences such as *Jack and Jill fetched water and wine, respectively*. Tree graph models for such sentences need to have the same number of branches in two places in a derivation, $a^n$ ... $b^n$, motivating the term *indexed grammar* for strings with these properties. If there is no limit on the index $n$, then clearly the finite working memory constraint is violated, independently of any type of embedding. However, there are small finite limits on $n$ (Kindt 1998; Karlsson 2010). Cross-serial dependencies are discussed further in Section 6.

Another case where more complex grammars may appear to be necessary are the *long distance dependencies* with *wh*-fronting, as in *Who$_x$ did John ask Mary to invite $\varepsilon_x$?* In this example, $\varepsilon$ marks the unfilled object slot which relates to *who*, and the pair *who$_x$* and $\varepsilon_x$ are in a long distance dependency relation between the filler and the object slot. In any given context there is a small finite number of such items, in many cases only one, depending on the argument slots in the relevant subordinate clause. Consequently, treatment as a regular language is not jeopardised. Gazdar et al. (1985) showed that these items can be described with a restricted generalisation of phrase structure grammars. If the phrase structure grammar is right branching, then *a fortiori* the relation can also be described by a regular grammar or finite state automaton. Section 6 contains more discussion of long distance dependencies.

The most complex language type in the Chomsky-Schützenberger hierarchy is the Type 0 or *unrestricted language*, described by an *unrestricted grammar*, modelled by an automaton known as a universal Turing machine, which can execute arbitrary string operations. The *transformations* of the first twenty years of Chomskyan grammars were of this type, and are now considered too general and too uninformative for describing specific insights into natural languages. Nontransformational descriptions for familiar relations which have traditionally been described by transformations are discussed in Section 6.

There is an inclusion relation connecting the languages in the formal language hierarchy:

$$L_{regular} \subset L_{context\text{-}free} \subset L_{context\text{-}sensitive} \subset L_{unrestricted}$$

Similarly, Type 3 grammars are special cases of Type 2 grammars, Type 2 grammars are special cases of Type 1 grammars, and Type 1 grammars are special cases of type 0 grammars. Types 0, 1 and 2 require unrestricted working memory and therefore do not fulfil the *Procedural Plausibility Condition* and Type 3 or regular grammars, and finite state automata, remain as the default or unmarked grammars in Multilinear Grammar. Marked non-default variants must be augmented with additional working memory which, however, also needs to be finite. There are many more grammar types and properties which a full account of the relevance of the Chomsky-Schützenberger hierarchy would provide but which are not directly relevant here; cf. Hopcroft et al. (2007).

## 2.4   A note on types of recursion

Recursion is a key concept, particularly at the phrasal rank. The main types of recursion are not always clearly distinguished in the linguistic literature, but can be identified as follows.

R1. *Recursion in general definitions of infinite sets of structures which can be represented by rooted tree graphs.* This recursion concept underlies the general 'merge' functions of recent generative linguistics which have been considered to be the core design feature of language (Hauser et al. 2002; Fitch et al. 2006); this concept has been discussed in some detail in the collections of van der Hulst (2010) and Roeper and Speas (2015). The epistemological point to make in this connection that this kind of recursion is an unlikely candidate for distinguishing between language and other domains of human experience, since pretty much any other domain from architecture to social structures and decision tree planning strategies can be analysed, at least in part, in these hierarchical terms.

R2. *Apparent recursion in strictly layered and other finite depth tree hierarchies.* If the hierarchy has finite maximum depth, there is no recursion back to a higher-ranking category, it can be processed by an acyclic automaton with finite working memory, and there is a maximum processing time constant. This condition applies not only to the prosodic hierarchy of Selkirk (1984) but also to SAAD (simple active affirmative declarative) 'kernel' sentences (Chomsky 1957), or the more general RSCs (root small clauses) of Progovac (2010), which are defined as simple clauses with no embedding of relative or complement clauses or phrases. Many grammatical constraints apply solely to root small clauses and therefore do not affect the procedural plausibility conditions. A caveat: in some linguistic descriptions, recursion is postulated because one node dominates another node of the same name in the finite environment of a root small clause. In such cases the node-name results from a stipulation, is simply structurally ambiguous, and does not provide a sufficient condition for recursion in the strict sense, which involves arbitrary repeated applications of a rule to its own outputs.

R3. *Recursion in purely head-recursive or purely tail-recursive grammars.* This type of recursion was discussed previously, and requires only linear processing time and finite working memory, thereby fulfilling the *Procedural Plausibility Condition*. Tail-recursive hierarchies are very common in linguistic descriptions, and indeed sometimes a condition is stipulated that all branching in a particular domain must be strictly right branching, at least in the unmarked default case: *this is the dog that chased the cat that worried the rat...* It is true that infinite recursion of any kind overgenerates the language: production and perception of arbitrary length strings are limited by arbitrary time and energy. Even though the plausibility condition is fulfilled, additional production time limits are required for a fully realistic grammar, just as they are required in domains as disparate as phonetic production or politeness in Grice's maxims of quantity and manner (Grice 1975).

R4. *Recursion over tree hierarchies, as permitted by general context-free grammars.* General context-free recursion is only possible at the expense of unbounded working memory and nonlinear processing time. This kind of recursion is by far the most popular in linguistics, and it clearly over-generates the language not only in terms of arbitrary string length but also in terms of arbitrary depth of embedding and thus also in terms of non-finite working memory. Restrictions on working memory size as well as on production time are needed, even in language registers with storage enhancement such as writing and rehearsed speech. Section 6 discusses these issues.

R5. *Tree hierarchies with cross-connections between the branches*. Unbounded and cross-serial dependencies can be dealt with using indexed grammars, a subset of context-sensitive grammars. This is also at the expense of unbounded working memory if arbitrary recursion is permitted. In this case, too, in order to satisfy the premise of plausibility and prevent infinite overgeneration the unbounded working memory requirement must also be tamed by specific finite working memory structures in order to handle the overgeneration problem. These issues are discussed in Section 6.

The first type, R1, is really a metatheoretical type which is applicable to life, the universe and everything. The others, R2, … R5, are types which are relevant for descriptive modelling, and types R2 and R3 may well represent the bottom end of a complexity development scale in the polygenetic and ontogenetic development of languages. This development scale may be extended to types R4 and R5 if appropriate additional finite working memory constraints on overgeneration are defined.

Since the Multilinear Grammar approach takes properties of processing time and working memory into account, an objection may be raised that the approach is about performance and not competence. This would be partly true: in Multilinear Grammar, well-defined procedural issues are systematically carved out from an otherwise vaguely defined area of performance and reassigned to the centre of linguistic concerns. Formal procedures are understood to be at least as important as static structures which are free of time and space constraints. Procedural plausibility is a necessary condition both on a realistic grammar and on 'working models' which operationalise a realistic grammar.

# 3  Characterisation of the Rank Interpretation Architecture

## 3.1  Background

Several aspects previously discussed from the perspective of Multilinear Grammar are taken up again and shown from the perspective of the Rank Interpretation Architecture. The Rank Interpretation Architecture, with its Peircean semiotic motivation and equally Peircean ternary dimensionality (Peirce 1905), has been influenced by several previous approaches to semiotics and linguistic theory, each of which incorporates some notion of 'level of abstraction', 'level of description', 'level of representation', 'stratum' or 'rank'. Precursors are found most clearly in the rank scale of Halliday's scale and category grammar (Halliday 1961), or in the levels of tagmemes in tagmemics (Pike 1967), or the concept of stratum in stratificational grammar (Lamb 1966), as well as in the more restricted rank-like concepts of the duality (Hockett 1958) or double articulation (Martinet 1960) of words in terms of morphology and phonology. The clause-rank concept of Jespersen (1924) is more related to phrasal linearity and hierarchy than to the present rank concept.

The architectures of these approaches are very different, deriving as they do from different structuralist and functionalist approaches, and covering the linguistic domain to differing extents. The approaches share with the Rank Interpretation Architecture framework both the semiotic premise that sounds and meanings are related by signs and structures of language, and the premise that the architecture of languages covers signs of different ranks, sizes and time windows, from discourse events through utterance, phrase and word to morpheme and phoneme.

Nevertheless, the Rank Interpretation Architecture framework has a fundamentally different semiotic basis from previous frameworks. In particular, there is no assignment of phonetics to the 'bottom' of a hierarchy of components, with semantics or pragmatics at the 'top' and syntax in between. Nor is there an inclusion scale of *Syntax* ⊂ *Semantics* ⊂ *Pragmatics*. Nor are there assumptions about function determining form or form determining function. In the Rank Interpretation framework, each rank has its own linear default structuring with unique semantic-pragmatic and prosodic-phonetic interpretations and in addition requires the plausibility constraints of linear processing time and finite working memory space.

## 3.2 Summary of the Rank Interpretation Architecture

The ranks in the Rank Interpretation Architecture α (α for ancient Greek ἀρχιτέατων 'architect') are characterised compactly as an *emergent rank sextuple*, developing in the species or the individual from as simply structured *discourse* rank through more and more intricate rank structures of *utterance*, *phrase* and *word*, each with its own semiotic pair of semantic-pragmatic and prosodic-phonetic interpretations:

> α = < *discourse, utterance, phrase, word, morpheme, phoneme* >, or
>
> α = < $α_{disc}$, $α_{utt}$, $α_{phrase}$, $α_{word}$, $α_{morph}$, $α_{phon}$ >, or
>
> α = < $α_1$, $α_2$, $α_3$, $α_4$, $α_5$, $α_6$ >

The morpheme and phoneme ranks are sub-ranks of the word rank. The ranks are emergent in that they are the results of different communicative functionalities which have arisen in the course of evolution (polygenetic emergence) and arise in language acquisition (ontogenetic emergence).

Needless to say, the detailed properties of ranks vary typologically in different languages and language varieties, in which different and more or less detailed structures develop at each rank. Dialogue conventions, different text genre conventions, different phrasal grammars, morphologies and phonologies diverge in different instantiations of the Rank Interpretation Architecture. There may be interdependence, fuzzy borders and transitions between ranks, for example between utterance and phrase rank in expressing speech acts, between phrase and word rank in morphosyntax, morphological incorporation, or nominal phrase to noun rank-shifting as with *the King of Spain's daughter*. Similarly, there are fuzzy relations between complex words and the morpheme sub-rank. Examples can be found in the increasingly nontransparent lexicalisation of compounds into morphemes in historical time, such as the completely opaque *husband* or *hussy*, from Old English *hus+bonda*, master of a household, and Middle English *hus+e+wif*, mistress of a household, respectively. But the upper and lower ends of the rank scale are well-defined: discourse is a recognisable kind of interactive social event, and the speech sound or phoneme, whatever the theoretical details, is a minimal word-distinguishing event. Indeed a minimal utterance, or indeed a minimal discourse, may skip lower ranked items, down to single words, as in the exclamation "John!", or to solitary speech sounds, overriding prosody choices at lower ranks: "Oh!", "Mm.", "Ah!", "Sh!".

Each individual rank $α_i$ from the discourse rank through to the phonemic rank is characterised as a *semiotic triple* of structures, syntax or tactics, $τ_i$ (Greek τ 'tau' for τάξις 'taxis', structure or arrangement), a semantic-pragmatic interpretation $σ_i$ 'sigma' (σημαντικός 'semantikos' significant), and $φ_i$ ('phi' for φωνή 'phone', sound or voice): $α_i$, = < $τ_i$, $σ_i$, $φ_i$ >.

The units and their interpretations are *emergent* in the sense that they are functions of the semiotic relation between sign events and the contexts in which the sign events unfold on scales of utterance time, acquisition time, historical time or evolutionary time. Spelled out, the Rank Interpretation Architecture is multidimensional, a sextuple of ranks with the semiotic pair of semantic-pragmatic and prosodic-phonetic interpretations at each rank: << $τ_{disc}, σ_{disc}, φ_{disc}$ >, < $τ_{utt}, σ_{utt}, φ_{utt}$ >, < $τ_{phrase}, σ_{phrase}, φ_{phrase}$ >, < $τ_{word}, σ_{word}, φ_{word}$ >, < $τ_{morph}, σ_{morph}, φ_{morph}$ >, < $τ_{phon}, σ_{phon}, φ_{phon}$ >>. The ranks are ordered in a strictly layered hierarchy with finite depth, but their internal structure may be either cyclic or noncyclic.

## 3.3 Properties of each rank

In order to motivate the Rank Interpretation Architecture it is necessary to show that properties, either of structure or of prosodic-phonetic interpretation, are unique to each rank, the *Sui Generis Condition*.

The same applies to semantic-pragmatic interpretation, though we do not discuss this dimension explicitly. The ranks share the processing properties which were discussed in Section 2: unmarked default processing at each rank is serial, linear, incremental and efficient, requiring its own finite working memory, conforming to the *Procedural Plausibility Condition*. The patterns at each rank are regular languages defined by regular grammars and modelled by finite state automata. Finite enhancements of working memory at each rank are possible for specialised registers such as writing and rehearsed speech. The components of the Rank Interpretation Architecture are outlined in more detail as follows.

1. *Syntax, τ*: a structural frame which differs from rank to rank but has default linear properties at each rank $\alpha_i$, with linear sequences composed of elements of $\alpha_{i+1}$. A discourse consists of utterance elements, an utterance of phrasal elements, a phrase of word elements, a word of morpheme elements and a morpheme of phoneme elements. The default architecture of the rank hierarchy is determined by a strict layering hypothesis, meaning that there is no inter-rank recursion between the ranks themselves, and compositional rules at each rank $\alpha_i$ refer only to the types relevant for this rank (bearing in mind a limited amount of fuzzy rank-shifting). There may be intra-rank linear recursion, in the default case linear or iterative recursion, which makes the entire system linearly recursive at each level. More complex types of recursive property are inherited from general cognition and from artistic, technological and spiritual cultural interaction rather than being defining properties of language itself.

2. *Semantic-pragmatic interpretation, σ.* Each rank $\alpha_i$ has its own semantic interpretation $\sigma_i$, which is incremental and linearly structured in the unmarked default case, but not necessarily so in view of need to express the complexities of general cognition and cultural context beyond language itself. Semantic-pragmatic interpretation is situation dependent and indexical, including pragmatic interpretation with connotative meaning at each rank, such as indexical voice features, or speaker, register, dialect and language specific variants. Semantic-pragmatic interpretation has properties which are specific to each rank:

   1. $\sigma_{disc}$: discourse framing, adjacency pairs, dialogue acts (speech acts in context), turn-taking, genres such as debate and conversation, participant role;
   2. $\sigma_{utt}$: speech acts; information structure; argumentation, narration, poetry and other text types;
   3. $\sigma_{phrase}$: propositional meaning, time, aspect, modality;
   4. $\sigma_{word}$: names, predicates and compositional operators;
   5. $\sigma_{morph}$: composed and simple names and predicates;
   6. $\sigma_{phon}$: *prima facie* a rather odd concept, but includes
      1. the contrastive encoding function of phonotactics, corresponding to the the 'duality' of Hockett (1958) or the 'dual articulation' of Martinet (1960);
      2. submorphemic onomatopaeic, cross-sensory and phonaesthetic interpretation, as described by Zelinski-Wibbelt (1983).

3. *Prosodic-phonetic interpretation, φ.* The parallel information channels of prosody and conversational gesture convey information which overlaps and is complementary to the locutionary channel of words, sentences, utterances and discourse sequences. The *metalocutionary hypothesis* (Gibbon 1976, 1981b, 1983) describes prosody as a parallel channel manifesting the semiotic functionalities of index and icon (Peirce 1905), with prosodic events such as pitch accents, global contours and boundary tones denoting, in the strict sense of the term, positions and simultaneous events in co-expressive locutions at each rank. The same considerations apply to conversational gesture (Gibbon 2011). Ohala's proposal of a universal sound-symbolic Frequency Code (Ohala 1994), with generally valid and personally and socially relevant interpretations of lower and higher frequencies, is a specific case of an approach to paralinguistic prosody as a parallel metalocutionary information channel. The metalocutionary functions of prosody are the indexical and iconic marking of the structure of speech; these functions hold at all ranks. The boundary tones and pitch accents of intonation are *metadeictic* markers of positions and degrees of relevance at the different ranks, a special case of metalocutionary functions (Gibbon 1983). Lexical tones and pitch accents, both phonemic and morphemic, augment the basic indexical and iconic functionality of prosody by introducing semantically arbitrary phoneme-like symbols. Prosody, particularly pitch patterning, plays a central role in the empirical underpinning of the Multilinear Grammar and Rank Interpretation framework. Prosodic complexification occurs in parallel with locutionary complexification in language phylogeny and ontogeny. Each rank $\alpha_i$ has its own unique prosodic-phonetic interpretation $\varphi_i$, ultimately observable in external physical terms:

1. $\varphi_{disc}$: intonation, rhythm, and vocalisations of discourse patterns marking the semantic interpretations at $\sigma_{disc}$;

2. $\varphi_{utt}$: pitch patterns and timing (including pause and deceleration) marking the semantic interpretations at $\sigma disc$; post-nuclear relaxation;

3. $\varphi_{phrase}$: intonation boundary tones, global contours and local pitch contours and durations associated with stress positions and pitch accents;

4. $\varphi_{word}$: morphophonology of inflection (including inflectional tone) and supralexical morphophonology such as assimilations and reductions between words, or tonal sandhi: cohesion marking by phonostylistic reduction, cf. *in Manchester* as [ɪm ˈmæntʃɛstə], *in Greenwich* [ɪŋ ˈgrɛnɪtʃ], *visionary* [ˈvɪʒn̩rɪ];

5. $\varphi_{morph}$: typologically very variable, with compounding and, in affixing languages, derivational word formation, including linking tones and stress position patterns, stress and pitch accent position, also including inflectional and word compositional tones in Niger-Congo languages such as Ibibio (cf. *ènò* gift, *àbàsì* lord, each with low tones, but with a linking high tone in *ènòábàsì*, a female personal compound name, similar to *Dorothy*, gift of God, a similar function to that of the linking infix *s* in German *Liebesbrief*, cf. Section 7);

6. $\varphi_{phon}$: distinctive features, including morphemic and phonemic lexical tones, stress positions and pitch accents.

Semantic-pragmatic interpretation $\sigma_i$ is not restricted by the procedural plausibility constraint of incremental and linear processing, though the finite working memory limitation also applies where the language structures and the prosodic-phonetic interpretation are concerned. The whole range of complex cognitive capabilities and cultural background knowledge, from art appreciation to the cross-referencing anaphora of variables in formal logic must be accessible to semantic-pragmatic interpretation, and must be reconstructed from the linear organisation of language structures together with available context.

## 3.4 Procedural perspectives on the rank hierarchy

Prosodic-phonetic interpretation has been shown in prosodic phonologies to require parallel processing, on the one hand of locutionary syllable, word, and sentence, utterance and discourse signs as *prosody-bearing units* (PBUs), and on the other hand of prosodic signs (tone, accent, intonation) as quasi-independent categories and structures. The prosodic-phonetic interpretations at each rank are related to the prosodic hierarchy, which deals mainly with structures at the ranks of word and sentence (sometimes referred to there as 'utterance'). The Rank Interpretation Architecture framework goes beyond sentence rank to include prosodic-phonetic interpretation in the larger time windows of the utterance and discourse ranks with explicit physical grounding in phonetics at each rank.

Prosodic-phonetic interpretation also models the parallel gestural activities of speech and their acoustic properties. Prosodic-phonetic interpretation is a special case of a more inclusive *multimodal interpretation*, which includes other gestural interpretations including writing, signing and conversational gesture (Gibbon 2011; Rossini 2012). Prosodic-phonetic interpretation also includes specifically temporal patterning: the complex epistemology of rhythm as an epiphenomenon deriving from various observational and cognitively constructed factors, the timeless 'categorial time' of phonology, the 'clock time' of digital phonetics, and the 'cloud time' of analog signals (Gibbon 1992; Carson-Berndsen 1998). The extensive literature on the roles of the rhythm and timing subdomain of prosody is not addressed in the present study (cf. Gibbon 2006; Wagner 2008; Gibbon and Yu 2015); the focus is on selected rank-specific properties of pitch contours in relation to the different time scales associated with the ranks of the Rank Interpretation Architecture.

Christensen and Chater (2016) propose three time scales in the context of language development: the phylogeny, ontogeny and utterance time scales. On the utterance time scale, Tillmann & Mansell (1980) distinguished further between three time scales for windows in prosodic processing which relate to subdomains of the rank hierarchy:

1. *A-Prosodie*: Suprasyllabic intonation, timing (including pauses), intensity variation;
2. *B-Prosodie*: Syllabic rhythms of the language;
3. *C-Prosodie*: The allophonic relations between phones;

As in Christensen and Chater's approach, the architecture of language is taken here to be an outcome of discovery procedures for different time scales, but we distinguish four linguistically motivated time scales, rather than the three of Christensen and Chater:

1. the speech time scale $TS_{speech}$ with short-term generation, tracking, analysis and understanding of improvised speech productions at different ranks in working memory;
2. the ontogenetic time scale $TS_{ont}$, with complexification of production, tracking, analysis and understanding as individual communicators mature;
3. the intergenerational historical time scale $TS_{hist}$, with modification of details of observation and reproduction as behaviours are passed from adult generations to child generations, with cross-links due to inter-language contacts;
4. the phylogenetic time scale $TS_{phyl}$, as the species evolves and the architecture of language complexifies.

In terms of combinatorial communication, the dynamics of the procedural time scales are empirically grounded in a diachronic implicational series:

$$TS_{phyl} \rightarrow TS_{hist} \rightarrow TS_{ont} \rightarrow TS_{speech}$$

Thus, if there is phylogenetic development, then there is historical development, if there is historical development then there is ontogenetic development, and if there is ontogenetic development then there is spontaneous development in speech processing as utterances unfold. The additional distinction between $TS_{phyl}$ and $TS_{hist}$ is made because there is a clear difference between two types of change:

1. historical changes over tens or hundreds of years, such as Grimm's Law (weakening of consonants in the emergence of the Germanic languages around 2500 years ago) or the syntactic change of *SOV* to *SVO* word order in the emergence of Middle English, coupled with loss of much inflectional morphology after the 11[th] century;
2. phylogenetic changes in the communication of the species over hundreds of thousands or indeed millions of years, such as the massive complexification of human communication in contrast to the vocalisations of nonhuman primate species, which have varied pitch patterns and rhythms but much simpler formant structure.

In the context of Multilinear Grammar it is relevant not only that the phonotactic systems of languages but also that many, perhaps all, sound laws in language change can be modelled by finite state transducers. This has been shown for Slavic languages by Kilbury (1997), Kilbury & Bontcheva (2004) and Kilbury et al. (2011). Grimm's Law, Verner's Law and the High German Soundshift can also be modelled fairly straightforwardly by finite state transducers. In fact, transducers for Grimm's Law and Verner's Law can be composed into a single transducer, showing complementary environments for the two laws, thereby indicating that the laws are intrinsically unordered, whatever their actual temporal order, and not necessarily ordered, as the traditional controversy would have it.

This was first shown by Kindt and Wirrer (1978) with their transducer-related separation of *Stellenzuordnung* (position assignment) and *Korrelationsbeziehung* (correlation connection). It is tempting to speculate that developments on the evolutionary time scale $TS_{phyl}$ are also subject to the same plausibility criteria of finite working memory and linear processing time as Kilbury's models on the historical time scale $TS_{hist}$.

We assume that gesture and prosody (in particular laryngeal gesture) play a major role on each of the four time scales. The beginnings of speech on $TS_{phyl}$ lie in simple rhythmic phase modulations and melodic frequency modulations of vocal and non-vocal gestures at the *discourse* rank, with communicative functionality, such as those found in the animal world. Increasing granularity of rhythm and melody develops as rank after rank emerges, from barely interactive discourses through unstructured utterances, then more complex rhythmic *phase modulation* reflecting headed structures, and melodic *fundamental frequency modulation* reflecting increasing complexity of sequences at phrase rank, culminating in the fine-grained *amplitude modulations* and *formant frequency modulations* of syllable patterning. We suggest that this developmental sequence can be observed most clearly on the language acquisition time scale $TS_{ont}$ (Speer and Kiwako 2009), and that an initial hypothesis for the phylogenetic time scale can be similarly characterised.

The analytic approach to developmental time scales, starting with simple patterns at the discourse level and increasing in granularity, relates better functionally and physically to empirical analysis than the prosody-free synthetic approach, with the bottom up accretion of phonemes to morphemes, and passage from one-word to *n*-word developmental phases. We therefore suggest that the Multilinear Grammar and Rank Interpretation Architecture is a coherent framework which makes more sense as a system of primary language design features than selected properties of sentences or arbitrary lists of essentially unrelated features. The *Procedural Plausibility Condition* on grammars and the *Sui Generis Condition* on ranks provide challenging evaluation criteria for realistic linguistic descriptions.

# 4 Discourse rank

## 4.1 The primacy of discourse patterning

The primary rank and the starting point for the development of gestural spoken communication on all time scales is the discourse rank, $\alpha_{disc}$. The primacy of discourse, its structures $\tau_{disc}$, its semantic-pragmatic interpretation $\sigma_{disc}$, and its prosodic-phonetic interpretation $\varphi_{disc}$, is simultaneously a hypothesis about evolution (phylogeny of language) and first language acquisition (ontogeny of language), as outlined in Section 3. Some sound is first uttered meaningfully in a discourse context, however short and unstructured. The increasing granularity of the utterance, phrase and word ranks develops as speech complexifies with time, with sounds expressing speech acts at utterance rank. The greater granularity of propositions follows at phrase rank with linearly structured sentences or sentence-like fragments, and with words expressing basic components of propositions. Increasing vocabulary enforces the development of linearly structured encoding of a very large number of morphemes by a very small number of phonemes in the word subranks of morphology and phonology.

It is these properties which cohere as Multilinear Grammar in the Rank Interpretation Architecture framework, and which represent the design features of gestural, spoken and written communication. In this section, unique properties of the discourse rank $\alpha_{disc}$ are demonstrated: default flat linear structural properties $\tau_{disc}$ and prosodic-phonetic interpretation properties $\varphi_{disc}$. Semantic-pragmatic interpretation $\sigma_{disc}$ is not discussed so explicitly.

The term 'discourse' has many definitions, ranging from communicative interaction to single extensive texts, such as Decartes' *Discours de la Méthode*. We hold a broad definition of discourse as *utterance in context*, which includes not only complex dialogues but also monologue in authentic pseudo-dialogue contexts (as opposed to cited monologues in isolation), including minimal or incomplete discourses. Examples are single-speaker utterances in contexts which may consist of a possibly unanswered call such as "Johnny?!", an interjection such as "Ouch!", or single-phoneme utterances such as "Oh!" or "Oh?", "Mm!" or "Mm?", with relevant prosody. The key feature of the discourse rank is the involvement of context beyond the speaker: a targeted interlocutor, whether perceived or not (as in unanswered calls or 'self-talk'), or an event which elicits appraisive interjections or a rhetorical monologue.

Models of discourse patterning $\tau_{disc}$ vary from the basic dyadic 'Q&A' (question and answer) stimulus-response patterns of behaviourist linguistics, through adjacency pairs in ethnomethodological conversation analysis (Sacks et al. 1974) to dialogue act sequences (Bunt 2011) as more complex adjacency tuples. Many formal models of dialogue patterning have been developed for human-machine communication, many of these being based explicitly on finite state automata (Bachan 2011; Griffiths et al. 2015a). Adjacency pairs and their generalisation to adjacency tuples are straightforward to model as regular languages. An early finite state approach (Gibbon 1985) included a number of types of loop which model question-answer-confirmation sequences and back-channel loops. Loops of this type can be compactly expressed as a regular expression $QA(C)(QA(C))^*$, where parentheses indicate optionality of the enclosed sequence and the Kleene star or iteration star '*' means that the item which precedes it, in this case ($QA(C)$), occurs zero or arbitrarily many times.

It is not impossible for dialogue sequences to be interpreted semantically at $\sigma_{disc}$ as having two or three ranks of 'discourse centre-embedding', as Christensen and Chater (2016) point out [*indentation added in order to visualise the proposed embedding*]:

    C1: I would like to purchase some cheese, please.
    S2:     What kind of cheese would you like?
    C3:         Do you have any French cheeses?
    S3:         Yes, we have Camembert, Port Salut, and Rocquefort.
    C2:     I'll take some Camembert, then.
    S1: (*wraps up some Camembert for the customer*)

But a closer look shows that centre-embedding is not the only possible analysis of this sequence. The semantic-pragmatic interpretation of the discourse progression at $\sigma_{disc}$ shows successive linear incremental clarification of a general category by means of a series of increasingly more specific hyponyms: *some cheese → French cheeses → Camembert, Port Salut and Roquefort*, followed by a decision. Centre-embedding is not required for this analysis. This discourse structure $\tau_{disc}$ fulfils the *Sui Generis Condition* by virtue of its own characteristic structure of linearly unfolding interactive negotiation.

## 4.2  Prosodic-phonetic interpretation: an adjacency pair

In the domain of discourse phonetics $\varphi_{disc}$ there are numerous prosodic patterns of pitch and speech rate (tempo) which are not characteristic of prosody at other ranks but are used to mark $\tau_{disc}$ components of adjacency pairs and triplets, such as questions, answers, short back-channel responses, or calls, or vocalisations such as *I see, mhm, aha, yes* (Couper-Kuhlen 2015).

Figure 1 visualises one type of pitch assignment which $\varphi_{disc}$ provides, extracted from a formal BBC radio interview in the Aix-MARSEC corpus[2] (Auran et al. 2004, recording J0104G): *Martin, do you think that the best side won it in the end? - Yes, I think there's no question that Argentina deserved to win...* Figure 1 shows the Q&A pair (top), the question (middle) and the answer (bottom). Each part of Figure 1 has a superimposed quadratic regression[3] curve which approximates the global $\varphi_{disc}$ pitch contour which marks dialogue act patterning.

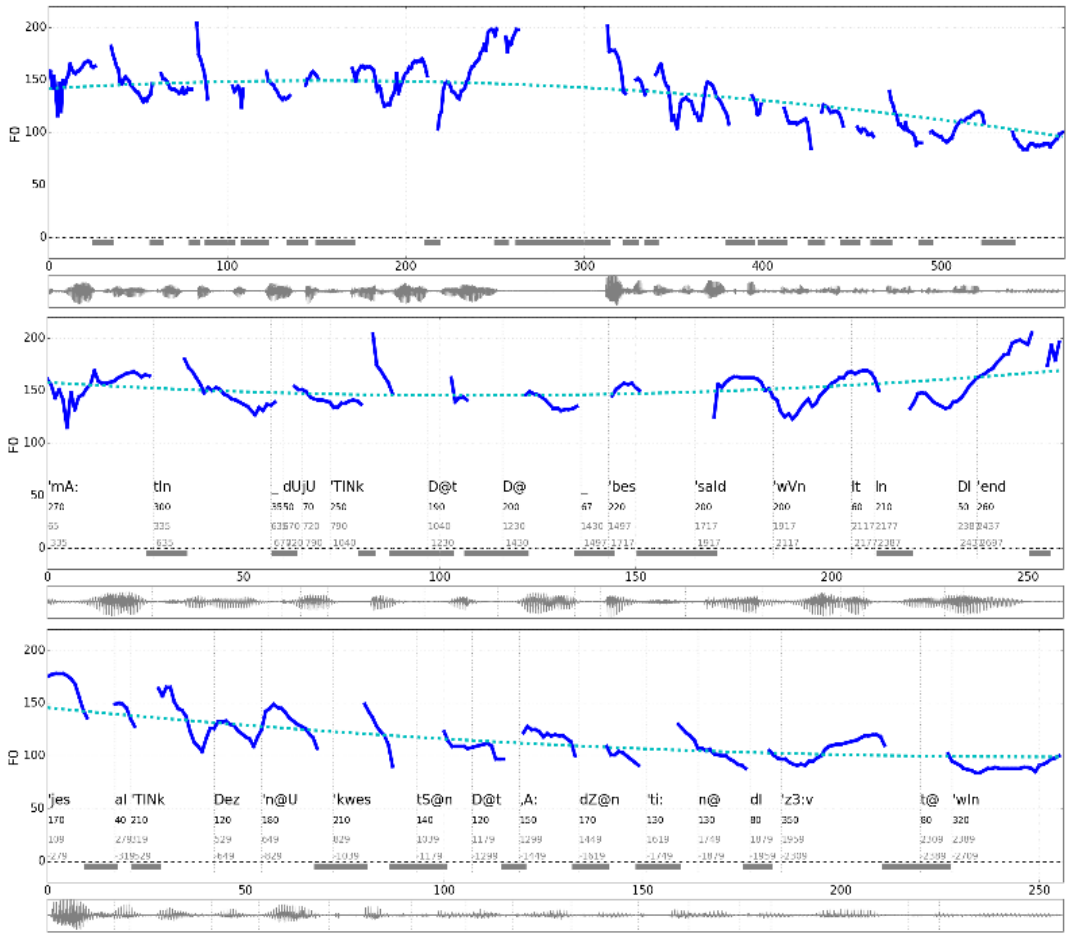

**Figure 1:** Question and answer (BBC interview, formal genre, male speakers, with quadratic regression lines.)

The relevant properties of the fundamental frequency contour are the following:

1. The average fundamental frequency of the question is higher than that of the answer. This is a tendency which has been reported elsewhere; cf. contributions to Cutler and Ladd (1983) and Hirst and di Cristo (1998).

2. The quadratic regression line is slightly rising for the question and falling for the answer, yielding an overall global rise-fall contour 'gestalt' over the adjacency pair.

3. Linear regression slope was also measured: 0.041 for the question (slightly rising) and -0.201 for the answer (a more pronounced fall).

4. Both question and answer terminate with a rising frequency modulation. The bandwidth (range) of the answer termination is much narrower than that of the preceding question. This points to unfinished or continuation status of the answer, which does in fact continue.

5. The pitch accents in the question (Figure 1, middle) are all low, on *MARtin, do you THINK that the BEST side WON it in the END?*; these pitch accent frequency modulations extend *below* the regression line. In contrast, in the answer (Figure 1, bottom) the pitch accents are all high, with excursions of the frequency modulation *above* the regression line. Depending on the particular language or language variety, this tendency can be associated with information structure: low pitch accents with old information, high pitch accents with new information.

The cooccurrence tendencies of the different fundamental frequency patterns observed in this example may be summarised as follows:

1. When a pitch accent shape is selected for the first accent position it tends to be retained for the following accents (with the possible exception of the final nuclear phrase accent).
2. Low pitch accents tend to co-occur with a flat or rising slope, high pitch accents tend to occur with falling slope.
3. Low pitch accents tend to be associated with introductory or known information, high pitch accents with completing or new information.

The metalocutionary function of the patterns is to make one or another segment of the dialogue more prominent by means of pitch movements, rhythm changes or breaks. The patterns described here are not unusual for English and have also been observed for other languages, as shown in contributions to Cutler and Ladd (1983) and Hirst and di Cristo (1998). The patterns are tendencies which depend dynamically on real-time incremental pragmatic decisions by the speaker on the speech time scale $TS_{speech}$. Default patterns may be overridden by higher ranked factors at any point and the cooccurrence tendencies observed in the $\varphi_{disc}$ mapping are clearly more intricate than is suggested by simple assignments of rising and falling terminal tones to question and answer pairs.

## 4.3  Chanted 'call' intonation

A prosodic feature which is exclusive to the discourse rank is the chanted stereotypic *chroma* feature of the so-called 'call contour'. The contour is not only used for calls but also for farewells, and, with more general metalocutionary functions, as a prosodic boundary marker which introduces or terminates uptake-securing phases of discourse (Searle 1969), or accompanies back channel discourse interruption markers of disagreement:

¯*John–ny!* ¯*Yoo–hoo!*
*Hel*¯*lo–o!* ... ¯*By–ee!*
¯*ʔmm–ʔmm!* (where 'ʔ' stands for the glottal stop.)

Aspects of chanted contours and their structure, function and phonetics have been discussed by several scholars, including Pike (1945), Liberman (1975), Gibbon (1976), Ladd (1980), Niebuhr (2013), Arvaniti et al. (2016) and Gibbon (2017).

Semantic-pragmatic interpretation $\sigma_{disc}$ has two main functions for this chanted contour in English (ignoring for present purposes related chanted patterns like the 'ritual insult' chants of children, liturgical chanting and rap):

1. conversational metalocutionary function, at normal volume, to start or terminate a conversation, for example to terminate a discourse on the phone or a shopping encounter;
2. teleglossic metalocutionary function (*teleglossia*: communication at a distance), in general with volume raised above normal conversational level, in order to attract someone's attention at a distance and to mark the potential beginning of a dialogue.

Examples of three chanted contours, recorded *ad hoc* in a citation context (not in an authentic calling scenario), are given in Figure 2 to illustrate *Hell*¯*o–o!*, *Goodb*¯*y–ye!*, and ¯*John–ny, where* ¯*are–you?* In an authentic discourse scenario both calls and responses may have the same chanted contour. It might be argued that these chanted contours are most appropriately dealt with at the utterance rank, and indeed the examples shown were not recorded in an authentic discourse context. However, the contours are discussed here in terms of their distribution as components of discourse, not in terms of the components of utterances.

The pitch traces show the normal irregular microprosody of vibrating vocal fold tissue: chanting of this type is not singing. The first case, *Hello*, shows a high pitch accent on the first part with a flat tail at mid pitch. The other cases have approximately flat pitch on both high and mid parts of the contour.

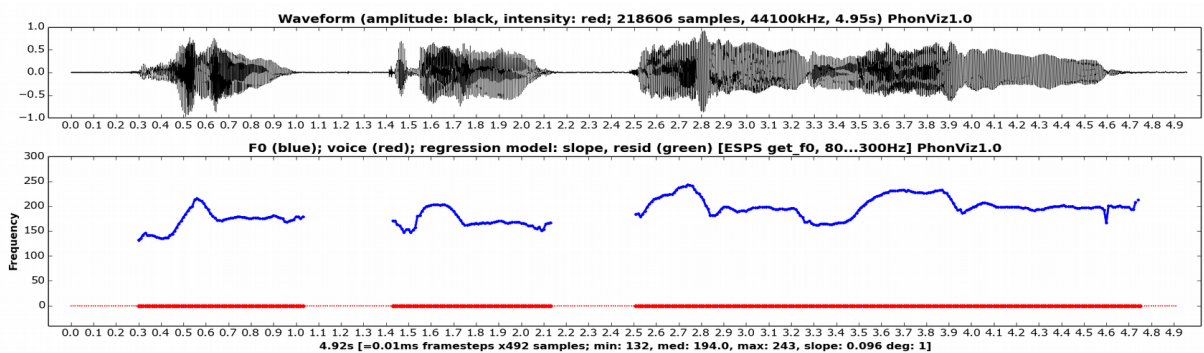

**Figure 2:** Waveforms and pitch traces of chanted contours (British English male voice) *Hello!*, *Goodbye!* and *Johnny, where are you*?

Table 2: Chant contour frequencies averaged over syllables and chant contour ratios compared with just and tempered minor 3rd (F0: fundamental frequency).

| associated locution | 1st F0 level | 2nd F0 level | F0 ratio | musical minor 3rd ratios | |
|---|---|---|---|---|---|
| | | | | just ratio | tempered ratio |
| *hello* | 212 | 177 | 1.198 | | |
| *goodbye* | 201 | 168 | 1.196 | 1.2 | 1.189 |
| *Johnny* | 240 | 196 | 1.224 | | |
| *where are you* | 230 | 197 | 1.168 | | |

It has been noted many times in the literature that the relation between the higher pitch accent and the lower tail is approximately a musical minor third (3 semitones, the 'lullaby interval'). The examples shown in Figure 2 and the syllable-averaged pitch values and ratios summarised in Table 2 illustrate the point.

It is important for checking the *Sui Generis Condition* on ranks that the chant contour relates to discourse rank $\varphi_{disc}$ only and does not co-occur with intonation patterns within sequences at the utterance or phrase ranks. The expression *I saw ⁻John–ny yesterday!* is quite odd, except perhaps for jocularly attracting the attention of Johnny across the room.

The basic chanted contour may appear *prima facie* to be a substantive universal of speech, but there are language-specific differences, and informal enquiries show that in some languages that only non-chanted contours occur in these discourse scenarios. In English, a rising version with an interval of approximately a musical major sixth (9 semitones) may also occur. In German, on the other hand, this does not happen (Gibbon 1976), but the functionality is broader: in addition to discourse initiation and termination functions, the chanted contour may signal an interruption of the discourse for a repair when securing of uptake for a successful speech act fails:

⁻*Lau–ter!* Louder! (To a lecturer.)

"⁻*End–lich!*" *hab ich gesagt.* "Finally!" I said. (A repetition after a misunderstanding, with flat pitch continuing to end of utterance.)

The differences between the discourse distributions of the chanted contour in English and German at the $\varphi_{disc}$ rank of prosodic-phonetic interpretation are modelled diagrammatically in Figure 3. Semantic-pragmatic interpretation $\sigma_{disc}$ for the chanted contour is schematically shown in the figure: the metalocutionary denotation of a particular position in discourse opening, uptake repair or closing.

The components of the discourse rank consequently fulfil the *Sui Generis Condition* and are structurally rather simple, also fulfilling the *Procedural Plausibility Condition* of linearity. For further information on discourse oriented intonation studies cf. Gibbon and Richter (1984), Selting (1995), Couper-Kuhlen (2015), Gibbon (2017).

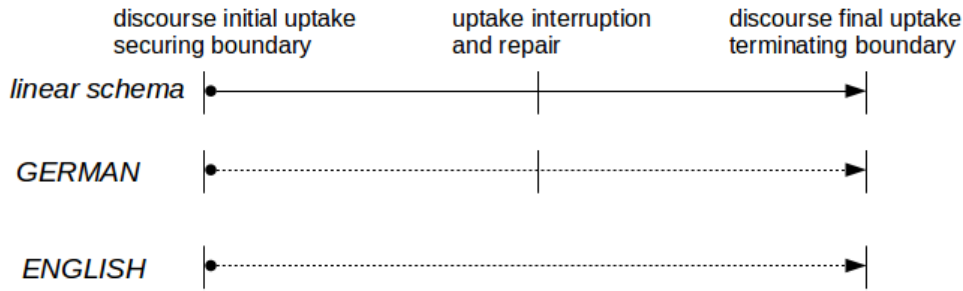

**Figure 3:** Linear schema for uptake establishing, repair and termination with chanted intonation contours.

# 5 Utterance rank

The specific compositional properties of the utterance rank $\alpha_{utt}$ result from the sequencing of sentences and sentence-like components by a single speaker. The structure of spoken utterances can be very simple or very complex, and is essentially semantically determined: non-iconic temporal ordering of sentences, inference patterns, narrative flashbacks, complex anaphoric and cataphoric cross-references, citations, speech acts, justifications, asides, quoting of entire narratives and discourses. The complexity of utterances is very genre-specific. On the one hand, in unpractised speech and some some forms of writing, the flat linear 'Hemingway style' of short sentences and linear structures is the unmarked default pattern, and more complex structures such as parenthetic or other embeddings tend to be highly stereotypic, for example: *by the way*, *I meant to say*, *if you will permit*. On the other hand, the extended working memory resources of rehearsed utterances and read-aloud written texts permit much more flexibility in types of recursion than the types which available for spontaneous speech: scholarly and legal texts are prime examples. The discourse context for rehearsed utterances and written texts offers the luxury of additional time and working memory resources, albeit with limitations, in both production and perception. The additional resources are supported by familiarity induced by repetition, lexicalisation, rehearsal and learning, and by the use of paper and screen as extensions of working memory.

Particularly important is what happens in prosodic-phonetic interpretation $\varphi_{utt}$. Utterance rank prosodic-phonetic interpretation most obviously involves pauses and long-term pitch patterning. Figure 4 shows a news item from the Aix-MARSEC corpus, read by a female BBC newsreader (Auran et al. 2004, recording A0101B), which illustrates some properties of $\varphi_{utt}$.

The point of using such a wide temporally compressed window for the pitch contour of a rather long utterance (just under 60 seconds) is to show the utterance level prosodic characteristics $\varphi_{utt}$ without distraction by lower ranked details.[4] There is no space for readable transcriptions of the locutions, but brief synopses are given below where needed. The most obvious structuring elements in the utterance are the pauses; the nine silent intervals which are longer than 200ms are numbered; these pauses precede larger and smaller *pitch reset* events (Gibbon 1981a).

---

4  A custom visualisation tool, *PV* (*Prosody Visualiser*) was developed for the purpose of visualising long pitch traces which cannot be displayed easily with popular software tools (cf. http://wwwhomes.uni-bielefeld.de/gibbon/PV/).

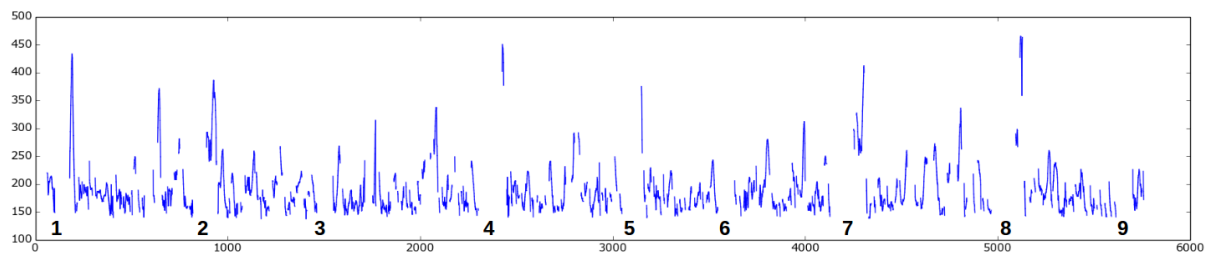

**Figure 4:** Pitch track (just under 60 seconds) of part of a news item in a BBC news broadcast, read by a female newsreader, from the Aix-MARSEC corpus (Auran et al. 2004). Pauses are numbered 1...9.

An initial greeting, *Good morning*, precedes *pause₁* and has its own falling pitch contour. In the interpausal unit between *pause₁* and *pause₂* the main topic of the news item is introduced. Between *pause₂* and *pause₃* the relevant event is described, and between *pause₃* and *pause₄* additional background information is given. A longer pause, *pause₄*, starts an interpausal unit with initial information for a new topic. Further information is added by the following interpausal intervals. The short final postpausal section shows a fragment of the next topic. In this highly formal variety of read speech the metalocutionary function of both pauses and pitch reset events in marking boundaries of narration segments at $\sigma_{utt}$ is fairly clear. In spontaneous speech the speed and bounded working memory constraints lead to greater variability than in the read speech illustrated in Figure 4 because of disfluencies and hesitations and rhetorical pauses.

A conspicuous property of $\varphi_{utt}$ prosody for this female speaker is the unusually extreme prominence boost associated with the pitch reset events introducing new information chunks. The pitch accent startup modulation pauses appear in Figure 4 as high upward frequency spikes of up to more than 400 Hz, with a following rapid drop in frequency (see also Figure 7 for a close-up of the first sentence in the sequence). This very high frequency on post-pausal pitch accents is not uncommon with female speakers of educated southern British English, particularly in formal registers. The frequency of boosted pitch peaks tends to decrease throughout the news item, with occasional minor resets which start new episodes within a news item. The highest reset peaks start a new news item: the height of the peak after pause 4, for example, exceeds the height of the first peak.

The boosted pitch reset events at the utterance rank provide evidence for paratone or paragraph intonation structure (Brazil et al. 1980; Lehiste 1975; Wichmann 2000; Tseng et al. 2005) with a longer prosodic time window than the single intonation unit. The data suggest that the paratone structure may be more complex than just two pitch heights, with pitch resets and pauses of different sizes depending on the speaker's formulation strategies (Bolinger 1972). The paratone sequence is reflected clearly in the fundamental frequency trace in Figure 4: each interpausal unit starts with an initial boundary effect of rising frequency on the anacrusis ('pre-head' of initial unstressed syllables) if present, then the very high topic-introducing pitch accent spike, followed by a sequence of less prominent pitch accents until the next spike. The spike-initiated sequences are terminated by a falling pitch accent and end in a 'tail' with post-nuclear relaxation, also called post-focus compression (Xu 2011), featuring post-nucleus reductions, post-nuclear destressing, de-accented, unstressed, reduced or deleted syllables, reduced frequency range, and slower tempo, including final syllable lengthening. There is no evidence of centre-embedded prosody in the paratone sequence, which therefore counts as fulfilling the *Procedural Plausibility Condition*.

The properties of regular pause distribution and structured sequences of interpausal pitch patterning, with paratone grouping indicated by highly marked pitch reset, are sufficient to show unique properties of utterance rank prosodic-phonetic interpretation, $\varphi_{utt}$, versus discourse intonation. It might be thought that the structure and phonetics of the utterance rank are simply an iteration of structures and phonetics of the phrasal rank, but the evidence provided by paratones in longer time windows refutes this conjecture and supports the unique character of the utterance rank. Text semantic analysis $\sigma_{utt}$ of utterances in terms of narrative and argument for is a further source of evidence for fulfilment of the *Sui Generis Condition* but is not discussed further in the present context.

# 6  Phrase rank

## 6.1  Characteristics of phrasal structure

Phrasal structure $\tau_{phrase}$ at rank $\alpha_{phrase}$, involves the combinatorics of components of phrases, clauses and sentences, and has properties which are unique to the phrase rank. The phrase rank has arguably been investigated more rigorously than any other in linguistics and it is more difficult to demonstrate the linear structural properties of $\alpha_{phrase}$ than of other ranks. Phrasal structures are different from both higher ranked utterance patterns and lower ranked word structures, though the boundaries between utterance structure and asyndetic phrase sequences on the one hand, and between phrase and rank-shifted or incorporated words on the other, may sometimes appear to be unclear. Controversy has arisen around centre-embedded recursive structures, which require non-finite working memory and processing in polynomial time. Centre-embedded structures may require exponential time with some backtracking algorithms for handling ambiguities, as in garden path sentences: *the horse raced past the barn collapsed*. One of the motivations for centre-embedding is facilitation of analysis of structural ambiguity and synonymy, which are, however, semantic rather than strictly combinatorial. The previously invoked thought experiment of listening to a lecture with unknown vocabulary illustrates the need for strict separation of semantic-pragmatic interpretation from plain combinatorial syntax.

It has been pointed out many times that empirically observable prosodic grouping and stress position levels are too shallow for the grammatical structures predicted by generative grammars, most notably by Chomsky (1965), Bierwisch (1966), Reich (1969), Langendoen (1975), Culicover and Rochemont (1983), Nespor and Vogel (1983), and in many later studies. The prosodic mismatch has either led to the consignment of these prosodic properties to an ill-defined domain of performance or was handled by structural depth-reduction and linearisation using restructuring or readjustment rules.

Non-finite working memory and polynomial or exponential time complexity may indeed be convenient heuristic assumptions for some purposes, but are quite implausible as empirical criteria for a well-founded realistic grammar. Productions which require such strategies provoke failures. Karlsson (2010) presents evidence for strong constraints on depth of recursion, limiting it to finite depth and thus to tractability with finite working memory systems. Phrasal patterns and their properties have mainly been discussed in terms of

1. rooted tree graph models which appear to require context-free grammars,
2. constraints which cut across tree graph structure, as in cross-serial dependencies, which appear to require context-sensitive grammars.

Some of these discussions are more relevant to semantic interpretation than to plain combinatorial syntax and to written registers and rehearsed speech. In unrehearsed informal spontaneous English speech in everyday contexts, nested self-embedded constructions are rare, partly because utterances tend to be rather short. An apparent exception is found in stereotypic parentheses, often with *verba dicendi, cognoscendi et sentiendi,* such as *I would say*, *if I may make a suggestion*, *and I am quite sure about this*, or *I feel embarrassed to say*.

An empirically observed example of the failure of centre-embedding in spontaneous speech occurs in the CHRISTINE1 corpus of informal spoken English (Sampson 1996). This corpus of 10394 sentences is a random extract from the demographically sampled British National Corpus. A search was made with clause-initial *who/whose/whom* as the search criterion. Other types of relative clause, e.g. with initial *that*, *which* and zero pronoun, were not investigated. The form *whom* did not occur. The search failed to find any successful cases of nested constructions: 145 sentences (1.4%) contained *who/whose* pronoun occurrences, of which 129 (1.24%) were sentence-initial interrogatives. Of the 16 sentences (0.15%) which were relative *who/whose* clauses, 9 (0.09%) were interrupted fragments. Only 7 (0.07%) were complete relative clauses. All complete relative clauses were right branching; none was centre-embedded.

In the majority of cases of *who/whose*-initiated relative clauses, an attempt to recurse was broken off. In the only clear example of a potentially centre-embedded *who/whose* relative clause, the sentence petered out incohesively and without completion of the top embedded *that*-complement [*formatted to show embedding*]:

> we found out
>> that the neighbours on the left hand side
>>> who were in fact an elderly couple
>>> and his was erm
>>> and he had his own business working at home
>
> [ End. The expected predicate phrase for the *that*-clause is missing ]

The continuation after *who* actually ignores *who* and the re-structuring is signalled by the hesitation *erm* and the self-repair in the last sentence. If, as we have argued, spontaneous speech is linear in the unmarked default case, and processing of nested sentences requires more working memory and processing time than is spontaneously accessible, then this failure is expected.

## 6.2   Linear sequences, iteration and regular grammars

Kernel sentences or root small clauses are describable by means of acyclic regular grammars. Juarros-Daussà (2010) notes that such constructions have a maximum length of 4 obligatory items, *Subject-Verb-Object₁-Object₂*, as in *Jack gave Mary Camembert*, conforming to a *Two Argument Restriction* on post-verbal complements. The limit can be flexibly, but still finitely, extended by optional adverbial constructions. Discussion of the increasing value of number of arguments (so-called one word, two word, *n*-word utterances) has figured prominently for many decades in studies of first language acquisition, and more recently in discussions of language evolution: from $n = 0$ (silence is no doubt an also an option for communication), through $n = 1$ for single word utterances, $n = 2$ for binary, and $n > 3$ for the general case. Jackendoff and Wittenberg (2016) suggest that the move from sequences of 2 to sequences of 3 was a qualitative leap in language evolution. Qualitative leaps are also represented by each move beyond sequences of 3, followed by unidirectional recursion and then by the centre-embedding generalisation.

Root small clauses are finite in length, and *per se* require only acyclic regular grammars. Internally, the fillers of argument slots in root small clauses may be cyclic but this does not change the finite length of the argument category sequence. The acyclic regular grammars conform tendentially to certain constraints documented by Karlsson (2010), at least in Germanic languages such as Swedish, German and English ('≺' signifies precedence): grammatical subject ≺ main verb; main verb ≺ grammatical object; main verb ≺ subcategorised locative; determiner ≺ nominal head; determiner ≺ adjectival modifier; adjectival modifier ≺ nominal head; case-marked genitival NP ≺ nominal head.

Domains of relations in root small clauses such as agreement relations (congruence, concord) are often described by means of context-free notations, but they can also be handled by acyclic regular grammars and are not context-sensitive in the strict sense. Agreement relations are based on small finite sets of features which can be represented in an automaton model as a finite set of storage registers. These storage registers are comparable with the storage registers in Augmented Transition Network (ATN) grammars (Wanner and Maratsos 1978; Gibbon and Eikmeyer 1983) and with the attribute-value pairs of the successor of ATNs, LFGL (Lexical Functional Grammar, Kaplan and Bresnan 1982; Bresnan 2001).

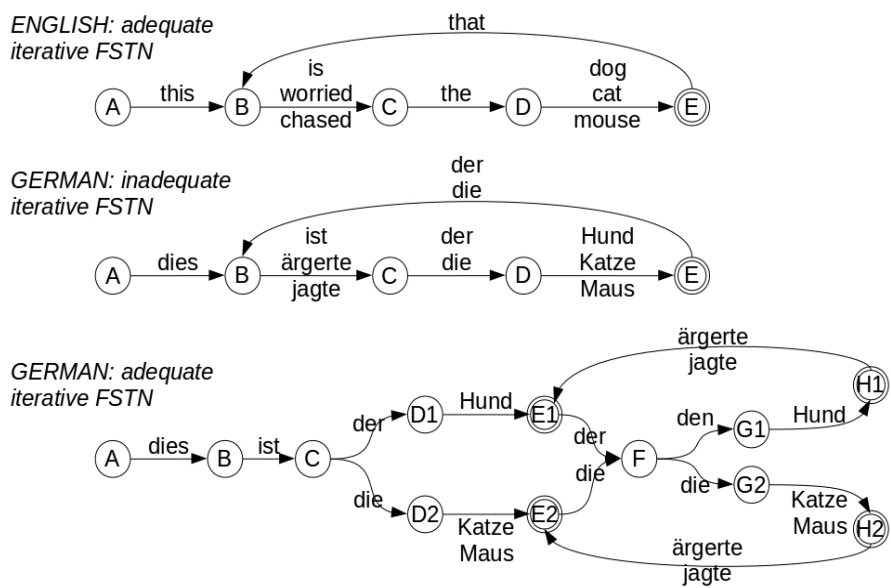

**Figure 5:** Finite state solutions for subject-verb congruence in English and German.

If the agreement relations are associated with right-branching structures, then a cyclic regular grammar provides a procedurally plausible solution. For example, intra-nominal agreement in German requires the feature set {CASE, GENDER, NUMBER} and nominal-verbal agreement requires the feature set {PERSON, NUMBER}. Agreement may occur across linear recursive contexts with relative pronouns, which initially looks problematic for the linearity condition, but if the construction is right branching, then the *Procedural Plausibility Condition* is still fulfilled. Furthermore, a finite state automaton with a finite set of storage registers may optionally be compiled out into an equivalent finite state automaton with no storage registers, such as the automaton illustrated in Figure 5.

Chomsky's auxiliary transformation[5] (flipflop rule, affix-hopping) for describing the dependency of auxiliary verb suffixes on the previous auxiliary verb in English is a powerful operation which is not necessary from a combinatorial point of view but reflects semantic grouping. An acyclic regular grammar is sufficient to describe the basic facts since the set of English auxiliary verb sequences is finite: $6 \times 2^4 = 96$ Tense-Aspect-Modality (TAM) combinations: five main modals (optional, i.e. six options), the obligatory present or past TENSE option, and optional PERFECT, PROGRESSIVE, and PASSIVE (i.e. four binary options). There is a maximal overt length of four auxiliary items, with tense always attached to the first verb, as in *it migh+t have be+en be+ing repair+ed*. The stem-suffix dependencies can be expressed by means of a linear automaton model in several ways: with 1-place lookahead, or with a register for each auxiliary, or simply with a finite state automaton (cf. Figure 6). Berwick and Pilato (1987) provide a similar analysis.

---

5  Let *Af* stand for any of the affixes *past*, *Ø*, *en*, *ing*. Let *v* stand for any *M* or *V*, or *have* or *be* (i.e. for any non-affix in the phrase *Verb*). Then: *Af + v → v + Af #* , where # is interpreted as a word boundary (Chomsky 1957:39).

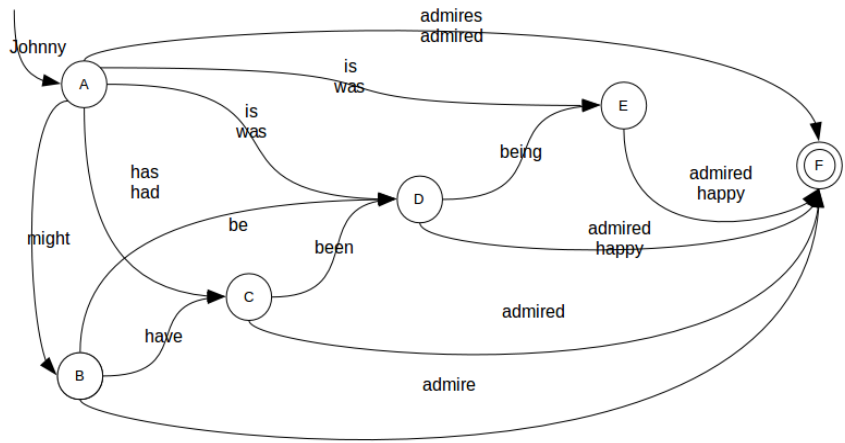

**Figure 6:** Finite state transition diagramme for English auxiliary verb affixation dependencies (restricted to one modal auxiliary option).

Karlsson (2010) also describes several constraints on branching constructions. One type is on left and right branching preferences, such as: 'an if-clause prefers initial embedding position' or 'a when-clause may be embedded initially or finally', both retaining the regular grammar property of left or right branching. A second type is a set of six distinct types of iteration in English (which are adequately modelled by regular grammars [*examples ours unless otherwise indicated*]:

1. structural iteration (e.g. conjunction): *Bonnie and Clyde* (syndetic); *veni, vidi, vici – I came, I saw, I conquered* (asyndetic);

2. apposition: *Ludwig Mies van der Rohe, the Bauhaus architect, once remarked that it was easier to design a skyscraper than a good chair.* - Lohr, Steven. 1992. *In the hot seat.* Chicago Tribune 1992-09-13);

3. reduplication: *It's a long long way to Tipperary but my heart's right there* – pre-WWI Anglo-Irish music hall song;

4. repetition (disfluent iteration): *Yeah, it, it, it is it's it's it's it's good.* (Biber et al. 1999: 1055, cited by Karlsson 2010:49);

5. listing (in restricted lexical taxonomies, e.g. proper names): *Jake, Jock, Jack*; (asyndetic); *Fitzgerald, Dietrich and Lady Gaga* (syndetic);

6. succession: *Monday, Tuesday, Wednesday, Thursday, Friday, Sat'day night / On the bedpost overnight* (Lonnie Donegan. 1959. *Does your chewing gum lose its flavour,* Skiffle adaptation); o*ne, two, three ... GO!* (trigger for synchronising simultaneous competition start).

In an analysis of plain combinatorial syntax it turns out that a regular grammar, and *a fortiori* a finite state automaton model for the grammar, is adequate for speech at the phrase rank. This even applies to cases which are usually described with more complex mechanisms. The obvious apparent exception is centre-embedding: where centre-embedding occurs, additional working memory and time must be available, as in writing and in rehearsed speech, but this also has strict limitations. The striking case of linearity in the Pirahã language for which lack of recursion has been argued (Everett 2005; Futrell et al. 2016), slots very well into the Multilinear Grammar framework. Clearly, asyndetic iteration or serial-like verb constructions also conform to the *Procedural Plausibility Constraint*.

## 6.3   A note on long-distance and cross-serial dependencies

Long-distance dependencies, which typically occur with *wh*-fronting and *verba dicendi, sentiendi et cognoscendi*, were previously thought to need transformational power. Gazdar et al. (1985) demonstrated with Generalised Phrase Structure Grammar that long distance dependencies can be dealt with by context-free grammars which have been enhanced with, essentially, a finite number of variables, frequently just one: *Who$_x$ did you say that John thought Joe had told Mary he had seen $\varepsilon_x$?* ($\varepsilon$ marks the object position in the source clause, to which *who* is assigned). The number of variables depends on the argument structure of the source clause for *wh*-fronting and therefore has a strict finite limit. A more complex case, still with a finite number of variables, is *When$_x$, where$_y$ and how$_z$ did you say John spoke to Mary $\varepsilon_x$ $\varepsilon_y$ $\varepsilon_z$?* Typically, such structures are right branching, and can therefore be described with a regular grammar plus a finite set of storage registers for the pronominal reference variables, comparable with the finite set of storage registers used to model inflectional concord.

The term 'mildly context-sensitive' was coined by Joshi (1985) for cross-serial dependencies such as *Jake, Jock and Jack married June, Joan and Jane, respectively*, where elements of lists of equal length are semantically matched in the given order; cf. also Shieber (1985), Michaelis (1998). This kind of pattern involves semantic pairing of nodes across trees, resulting in trees in which two or more nodes are constrained to have the same number of branches, each of which is paired in the same order with a branch of the other node. Formally, the constraint is comparable with a constraint on embedded tables with equal length rows and equal length columns. A more complex case of cross-serial dependency is *Jake, Jock and Jack gave June, Joan and Jane chocolates, flowers and diamonds at Easter, Whitsuntide and Christmas, respectively*, subject to the same conditions of semantically matched ordering, but generalised to longer tuples than pairs, and to more than two matched positions. Other kinds of cross-serial relation are found in some languages.

Cross-serial dependency appears to be unconstrained by embedding and to have more to do with semantic-pragmatic interpretation than with plain syntax: *Jake and Jock, who are married to June and Joan, respectively, will be visiting us tomorrow*. Another example: *When Jake and Jock came in, they greeted June and Joan, respectively*, or *If John says he saw Jake and Jock he really means that he saw Jim and Jack, respectively*. Multiple use of *respectively* is possible but clumsy, as in *Jake and Jock, who are married to June and Joan, respectively, are British and Australian, respectively*. The constraint seems to be that *respectively* tends not to appear more than once: *Jake and Jock, who are married to June and Joan, are British and Australian, respectively*.

The power of indexed languages (Gazdar 1987) and counting or copy languages (Kallmeyer 2010a, 2010b) has been said to be needed for modelling cross-serial dependencies. These Type 1 languages are more powerful than context-free languages and *a fortiori* more powerful than regular languages. The counter-argument is that the copies required are not for lists of arbitrary length requiring Type 1 indexed languages with unbounded working memory, but for short lists with a small finite bound of two or three, as Karlsson has shown (2010). The copying or counting operation in itself is linear in the length of the list. Consequently, if there is no arbitrarily deep embedding, the influence of the operation on overall complexity is minimal and unlikely to be empirically measurable.

A similar argument for linear modelling of cross-serial dependencies was advanced by Kindt (1998), who points out that separating ID (hierarchical immediate dominance) and LP (linear precedence) relations as parallel constraints enables a regular grammar to be used, also noting that cross-serial dependencies are empirically bounded and that the failure of regular grammars to generate unbounded cross-serial dependencies is thus not a counter-argument.

Semantic bounds on the cross-serial constraint are not as clear as they are usually made out to be. There are legitimate constructions with potential multiple ambiguities in unequal lists, as in *Jake and Jock married June, Joan and Jane*. Indeed, even with *respectively* the constraint does not need to be heeded, as in *Jake, Jock and Jack married June and Joan, respectively*: in informal tests with linguistically naïve native speakers various interpretations of this expression were offered but the acceptability of the formulation was rarely called into question. It is possible, therefore, that *respectively* may also have a more general meaning of 'ordered sequence' as well as the canonical strict meaning of 'paired in exactly that order'. This goes against received linguistic wisdom, but is no less valid for that, and strengthens the argument that cross-serial dependency is a semantic-pragmatic constraint, not a syntactic constraint, and that syntactically such structures are less cross-serial than simply linear.

## 6.4 Prosodic-phonetic interpretation at the phrasal rank

There are many different ways of looking at prosodic-phonetic interpretation at phrase rank, $\varphi_{phrase}$. In generative parlance, $\varphi_{phrase}$ is, on the one hand, 'postlexical' and deals with morphophonemic alternations and stress position assignment beyond the domain of the word, within a prosodic hierarchy, and on the other hand deals with phrasal intonation.

Traditional studies of the prosodic-phonetic interpretation of phrases, intonation and rhythm, have often relied on the hermeneutic methodology of intuited intonation patterns, i.e. imagined data, as in the mainly pedagogical frameworks analysed by Gibbon (1976), rather than observed data. This does not make these data necessarily less valid than 'measurable data', but they are far from the intonation of spontaneous speech. A common example of imagined data which is valid in principle concerns the disambiguation of the scope of structural and semantic operators, as in *I did not go because I was tired*, disambiguated with punctuation and prosodic boundaries, as *I did not go, because I was tired* (i.e. I did not go), contrasting with *I did not go because I was tired but because I was bored* (i.e. I did go). But higher ranked utterance and discourse factors may disrupt this simple analysis.

Halliday (1967) introduced a convenient trio of terms covering the main properties of phrasal intonation, which is still popular in English language teaching, and which may be rephrased in terms of later terminologies: *tonality* (phrasing, i.e. the assignment of boundaries and global contours to utterances); *tonicity* (the placement of stress positions associated with nuclear or phrase accents, which can also be generalised to the placement of any pitch accent, whether nuclear or not), and *tone* (the shape of the local pitch contour on the nuclear accented syllable, e.g. rising, falling, or a combination of these).

In generative phonologies, tonality and tonicity, i.e. phrasing and accent placement, were modelled cyclically, by means of the Nuclear Stress Rule, NSR, of Chomsky and Halle (1968) and later variants. The function can be represented by an algorithm formulated in either top-down or bottom-up fashion. A top-down formulation is as follows:

1. Assign value 1 to the rightmost constituent and value 2 to all others.
2. For each child constituent, recursively assign the parent value to the rightmost constituent, and the previously assigned value plus one to the others.

Thus a structure such as $((big\ John)\ (saw\ (small\ Joan)))$ is coded with the values $big^3\ John^2\ saw^3\ small^4\ Joan^1$, where the numbers represent stress positions and values. The tree-coding function is bijective, and thus applies not only to generation but also to parsing (Gibbon 2006). But it is not the whole truth in relation to prosody: higher ranking utterance and discourse factors may override the pattern, and in authentic speech as opposed to citation forms, linear patterns are generally observed. At various times, the implicit prediction of unlimited grammatical depth by the NSR has been disputed on the grounds of empirical implausibility. Repair measures were introduced in the form of depth-limiting readjustment rules (Bierwisch 1966; Langendoen 1975; Culicover and Rochemont 1983). Nespor and Vogel (1983) further disputed the need for a readjustment component, but their $\varphi$-restructuring rules have the same linearisation function as readjustment rules. The linearity of the accent patterns associated with stress positions was noted in the finite state automaton model of Pierrehumbert (1980), later taken up by other researchers, e.g. Ladd (1996).

Figure 7 visualises the first sentence from the newsreading example shown in Figure 4. The structure is linear: the 'sentence' lacks a main verb, and consists of a single extensive nominal phrase, which includes two linear recursions separated by a pause, each of depth 1: an appositional parenthesis, and a right branching relative clause. The example illustrates one kind of sentence intonation and how the linear structure of the sentence tends to be reflected by pause duration, with a shorter pause before the apposition than before the relative clause. There is no main verb, and syntactic structure is not closely reflected by the actual pitch. The very high paratone peaks mark the start of information patterns, the first of which coincides with a nominal expression with a prepositional modifier (*about…*) and an apposition relating to the modifier (*founder…*), and the second of which is a relative clause, (*who is…*). The first paratone peak points to a topic or known information, the second to a comment or new information.

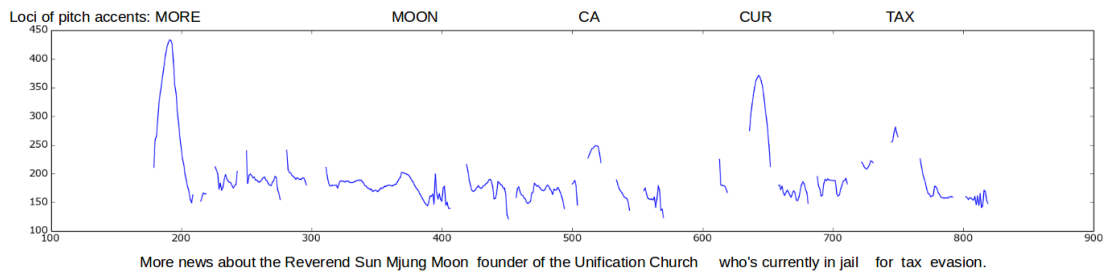

**Figure 7:** Pitch track for "More news about the Reverend Sun Mjung Moon, founder of the Unification Church, who's currently in gaol for tax evasion." (BBC female radio newsreader, Aix-MARSEC corpus A0101B.wav)

The example in Figure 7 visualises a number of characteristic features of sentence intonation in the newsreading genre. First a very high onset pitch accent (443Hz) occurs on the first lexical item 'more', which is characteristic of this type of speaker, as already noted in the context of $\varphi_{utt}$. The body of the first phrase has the expected declination and a final pitch accent on the name 'Moon' with final irregular pitch due to creaky voice (vocal fry).

After a short pause, a parenthetic recursion occurs on 'founder of the Unification Church': the pitch contour restarts slightly higher, with overall declination until the syllable *ca* of Unification, followed by post-nuclear relaxation with low pitch on the following short, reduced syllable. It is noteworthy that the pitch does not reflect the subordinated apposition, unlike canonical pronunciations in citation contexts. The pitch accent is considerably higher than the preceding accent on the antecedent noun *Moon*. The sentence continues with a right branching recursion on the relative clause *who's currently in jail for tax evasion*. A similar pattern to the preceding main clause occurs: high onset pitch accent modulation, though not as high as on the first onset pitch accent. Then there is pitch declination until the final nuclear pitch accent on *tax*.

The pitch pattern does not suggest a branching structure, but rather a spontaneously produced *ad hoc* iteration of linear patterns. The paratone grouping does not exactly match the sentence constituents, but attention is directed to successive iterations by pitch accents which the speaker assigns dynamically. The iteration sequences are marked prosodically by repetition of a global (declination) or a local (accent) pitch pattern and by similarity of timing (rhythm).

The phonetic form of the pitch accent is quite variable in a stress-based language like English: it can be high, low, rising, falling, rising-falling, falling-rising and there is currently no standard for distinguishing between representations of pitch accent shapes as contours (Halliday 1967 and in much of the pedagogical literature), as sequences of pitch levels (Pike 1945; Silverman et al. 1992; Pierrehumbert 1980), or as pitch target functions (Hirst 1998).

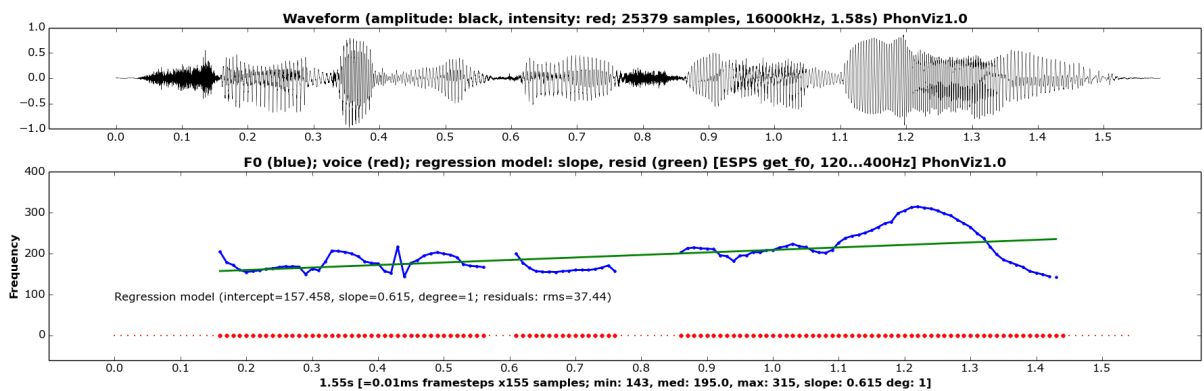

**Figure 8:** Waveform, pitch trace and voice trace of a BBC newscast extract (female voice, Aix-MARSEC corpus A0103B): "shared it with those who had NONE", with superimposed linear regression pitchslope line as illustration of positive slope (inclination rather than declination).

Figure 8 shows two patterns which already figured in the discussion of pitch patterns at discourse and utterance ranks:

1. Pitch accents inherit their shape from the first pitch accent in the sequence, up to but not necessarily including the nuclear accent.
2. If pitch accents are low, then slope is rising (the opposite tendency with high pitch accents may be observed when the slope is falling, as noted in connection with Figure 1).

Sequence constraints like these have often been listed in the pedagogical textbooks on intonation, under names such as 'stepping head', 'rising head', but in the linguistic literature, each pitch accent is generally modelled as unconstrained by its neighbours. In order to discover how general or how constrained these tendencies are, more empirical investigation is needed.

It is presumably uncontroversial that the structures of $\tau_{phrase}$ have properties which are unique to the phrase rank, fulfilling the *Sui Generis Condition*. There is strong evidence that the unmarked default grammar is linear in the sense outlined in the present study, and the linearity of prosodic-phonetic interpretation in $\varphi_{phrase}$ has been extensively documented in the literature and selectively illustrated here, underlining the validity of the *Procedural Plausibility Condition* at the $\alpha_{phrase}$ rank.

# 7 Word rank: morphemes, morpheme sequences, phonemes

## 7.1 Flat words

The word rank $\alpha_{word}$ is characterised by two sub-ranks: morphology is the 'upper half' of the duality (Hockett 1958) or dual articulation (Martinet 1960) of words; phonology is the 'lower half'. The subranks will be treated together with the word rank itself. The word rank is concerned with inflectional morphology, the morphological subrank with word formation, and the phonological subrank with phonological patterns.

The size of the inventory of syllables in the phonological subrank is finite. The reason is straightforward: the syllables of a given language are constructed out of a finite inventory of phoneme-sized constituents and have a maximum length (e.g. 9 in English, if diphthongs and affricates are counted as two constituents, otherwise 7, as in [ˈskraʊndʒd] *scrounged*). The number of potential syllables is therefore finite, of the order of 300 for Mandarin Chinese (without tones, and depending on the analysis criteria), and of the order of 30000 in English (Gibbon 2013). For a basic model of phonemic syllable phonotactics an acyclic regular grammar is sufficient (Gibbon 2007, 2013). In fact, Whorf (1940) already had a formula for syllable structures which is straightforwardly interpretable as a regular expression without iteration.

At the morphological subrank, the size of the lexicon itself is in principle unrestricted, like the set of sentences in the whole language. In the Germanic languages, compound word formation can famously be linearly extended to arbitrary lengths, at least in writing. A federal decree pertaining to Berlin, valid until 2007, illustrates the point in German; any number of whimsical further extensions are possible. Spaces are inserted to mark relevant structure and to highlight the linking *s*-morpheme:

> *Grundstücksverkehrsgenehmigungszuständigkeitsübertragungsverordnung*
>
> *Grund stück s verkehr s genehmigung s zuständigkeit s übertragung s verordnung*
>
> (*property transaction approval responsibility delegation decree*)

In fact this compound is an official abbreviation for a lengthier nominal phrase, and is further abbreviated officially as *GrundVZÜV*.[6]

One source of large but still finite sets of new words is blending, in which onset and nucleus components of attested lexemes such as *braid* and *flunk* are concatenated to form lexically unattested but possible syllables such as *brunk* and *flaid*. Blends in English are perhaps best known from the poem *Jabberwocky* by Lewis Carroll. The size limits on the total number of attested and innovated monosyllabic words of this kind in English can be defined precisely (depending on the analysis criteria) by multiplying the number of possible onsets by the number of possible nuclei:

$$|\text{Onset}| \times |\text{Nucleus}|, \text{ i.e. } 56 \times 549 = 30744$$

The innovated syllables are available for semanticisation and lexicalisation as neologisms, for instance in the creation of brand names such as *BiC*, *Twix*, *Skype*. The name *Stux*, for a state-created computer virus, is sometimes said to be derived from the ancient Greek Στυξ, Styx (the river dividing earth from hell, and meaning hatred), sometimes to be a blend, derived from *stinks and sucks*, and sometimes to be a geeky abbreviation for words starting with *stu* and with variable (*x*) standing for an unspecified ending, such as *stupid*, *stupendous*, *stuff* or *stun*. An unconventional blend, referring to graphical user interfaces (GUIs) for computers is *UX*, for *user experience*. Another relatively recent example is *twerk* (a kind of sexually provocative dance move, said to be a blend of *twist and jerk*). The innovated blend is not limited to monosyllables: *transistor* (*transfer resistor*), *Brexit* (*British exit*), *anecdata* (*anecdotal data*). Creation of these pairs is evidently a strictly linear operation.

One less conventional subdomain of the semantics of syllable constituents is phonaesthetics (Zelinsky-Wibbelt 1983). Phonaesthetic onset clusters perhaps originate historically in explicit compounds and blends, and fit nicely into the concept of a semantic-pragmatic interpretation $\sigma_{phon}$ of phonotactics. Examples are the highly restricted syllable onset phonaestheme *gl*, in *glance, glare, glass, glaze, gleam, glimmer, glimpse, glint, glisten, glitter, gloaming, gloom, gloss, glow, glum*, perhaps also *glade* and *glamour*, which are connected with aspects of sight or light. An equally highly restricted phonaestheme is *sl*, in *slake, slaver, sledge, sleet, slick, slide, slime, slink, slip, slither, slobber, sloop, slop, slough, sludge, slug, sluice, slurp, slush*, possibly also *slay* and *slew*, which are concerned with wet and slimy stuff. Phonaesthetic interpretations of onset clusters are rare, and phonaesthetic interpretation of these clusters is far from being exceptionless (cf. *glide* or *slim*).

For inflectional morphology the situation is clear: there is a finite set of inflection affixes, and a finite maximum length for combinations of inflections, yielding a finite set of inflected words for each lexical root. The discontinuity of circumfixes, as in German *ge-zähl-t* (*counted*) has finite length and a finite constituent inventory, and is therefore also unproblematic in terms of procedural plausibility. The finite set of affixes is quite small for languages like English, but very large for agglutinating languages like Finnish or Turkish. The applicability of finite state automata to morphological patterns, mainly for inflection, is well established in computational morphology, morphophonology, phonology and tonology (Johnson 1972; Koskenniemi 1983; Gibbon 1987a; Kay 1987; Kaplan and Kay 1994; Turhan 1996; Beesley and Karttunen 2003). For modelling alternations at morpheme boundaries a different kind of cyclical regular grammar was introduced by Koskenniemi in two-level morphology, in which the automaton cycles through a string searching for relevant contexts to apply the phonology-phonetic transduction.

---

6 http://www.buzer.de/gesetz/5188/index.htm

Striking cases of Multilinear Grammars at the morphological sub-rank are Kay's finite state transducer for Arabic which processes consonantal root templates which have vowel positions which are filled with parallel patterns of inflectional and derivational vowel sequences (Kay 1987). Equally striking at the phonological sub-rank is Kirchhoff's application of parallel stochastic linear automata (Hidden Markov Models) to phonotactic modelling in speech recognition (Kirchhoff 1996, 1999). Both these computational models are motivated by prosodic phonologies, particularly by autosegmental phonology (Goldsmith 1976, 1990).

In word formation (Marchand 1960) combinatorial patterns are also linear. Derivational and compound word formation are dealt with in the following two subsections.

## 7.2   Flat derivations

In principle the set of derivational suffixes is finite and the maximal length of a suffix sequence is in principle finite, with sequences like *regularisational*, and the set of derivational suffix combinations is also finite. However, in English derivational morphology there is a cyclic oddity: a sequence like *regional* can be extended to *regionalisation* and *regionalisational*. So in principle a further extension can be made, by induction on iteration: *regionalisationalisationalisational...* etc. The result is in practice strictly limited to one cycle.

Suffixal derivation is often described with left branching, reflecting part of speech conversion and the associated semantic interpretation: (A (N (V (V (A (N *region*) *al*) *is*) *at*) *ion*) *al*). For derivational prefixes the situation is a little simpler: in general they can occur in any order. That is, given a vocabulary such as $V_{pref}$ = {*over, post, pre, pro, trans, un*} the number of possible combinations is expressable as the simplest possible regular expression $V^*$. Whether a particular order or combination is currently meaningful is a different issue. For example: *antitranspostoverkill* may seem bizarre at first glance, but the bizarreness is semantic, not combinatorial, and it is nevertheless a possible derived word. The same applies to *overpostpretransantikill*, with prefixes in reverse order.

## 7.3   Flat compounds

If semantic-pragmatic interpretation is again kept separate from plain syntax combinatorics, the structural complexity associated with the compound words turns out to be semantic, being largely based on hyponymy and its effect on information structure. Compound nouns are thus easily described in linear terms. Given a vocabulary of simplex words $V$ = {*cam, cycle, cylinder, head, motor, over, shaft, twin*}, the vocabulary items can occur in any order as members of the set $V^*$, i.e. an infinite number of any combinations, including the following:

> *twin cylinder overhead camshaft motorcycle*
> *over twinshaft cam motorhead cycle cylinder*
> *camcycle cylinderhead motor overshaft twin*

A combination which makes sense to a biker, but perhaps not to many others, is *twin cylinder overhead camshaft motorcycle*. Other orders are possible and potentially meaningful. Nonsensical and near-nonsensical compound neologisms are regularly coined for new brand names or product genres, such as *Pepsi Cola*, *KitKat*, *Rice Krispies*, *DeskJet*, *WhatsApp*, *email*, *netbook*, *smartphone*, *hatchback* or created *ad hoc* as nonce words in particular situations: *that squishthingy*. Orthographic spacing conventions vary: in written English, spaces or hyphens in compounds, or lack of them, have a somewhat unsystematic relation to semantic hierarchies in $\sigma_{word}$ and to stress positioning in $\varphi_{word}$.

The semantic-pragmatic interpretation $\sigma_{word}$ of compounds has various complexities, unlike the linear combinatorics. For example, the word *motorcycle* is semantically an endocentric or *tatpurusa* compound (traditionally named with the Sanskrit term): the meaning of the whole is a hyponym of the meaning of the head or *determinatum*, often with additional lexicalised modification of the meaning and different semantic relations between the constituents. The semantic relation between the head and the whole is paradigmatic, i.e. classificatory. In English the determinatum of a tatpurusa compound is the right-hand element, modified by the meaning of the *determinans*, the left-hand element (Marchand 1960). In tatpurusa compounds, semantic interpretation of the compound in $\sigma_{word}$ is incremental and does not need right branching. In longer compounds such as *twin cylinder overhead camshaft motorcycle* the internal structure is semantically determined, while the combinatory properties are linear.

Bicentric (Sanskrit: *dvandva*) compounds such as *fighter-bomber* (cf. the lexically rare *bomber-fighter*), *creepy-crawly*, or *whisky-soda* (*soda-whisky* is also a possible compound) are similar to tatpurusa compounds. The difference between dvandva and tatpurusa compounds is that the meaning of the dvandva compound may be a hyponym of either of the two linear components.

In exocentric or *bahuvrihi* compounds such as *redbreast* (robin, i.e. a bird with a red breast) the semantic relation between the head (in this case *breast*) and the whole is syntagmatic, i.e. compositional: the constituents may have different parts of speech, and the structural head, *breast*, is not the semantic head, which is *bird*, and needs to be reconstructed from the compound in context. Similarly: *redhead* (someone with a 'red head'), *scarecrow* (something which scares a crow) or *pickpocket* (someone who picks pockets). The inverses *pocketpick* and *crowscare* are also possible bahuvrihi compounds, though *breastred* or *headred* may be interpreted as tatpurusa colour adjectives. Combinatorially, the bahuvrihi compounds are linear, like the root small clauses or kernel sentences they may be associated with.

## 7.4  Prosodic-phonetic interpretation at the word rank

The prosodic-phonetic interpretation of words $\varphi_{word}$ comprises the whole of conventional phonology and morphophonology, plus lexical prosody. This broad domain has been treated exhaustively in countless articles, monographs and textbooks, so we discuss selected aspects which have been less extensively treated in previous studies.

The treatment of phonotactics and the phonetic interpretation of phoneme variants in linear contexts by means of finite state transducers has already been noted. In morphophonemic typology, patterns of lexical stress positions in simplex, derived and inflected words are also standardly described in terms of linear position: initial (Hungarian), final (French citation forms), penultimate (Polish), or with complex dependencies on flat linear structures, as in English.

An apparent counterexample to the linear combinatorics of morphology and in particular $\varphi_{word}$, is found in the Compound Stress Rule (CSR) of Chomsky and Halle (Chomsky and Halle 1968) and its later variants, for compounding in English and related languages such as German. The CSR assigns stress patterns in compounds on the basis of both linear position and depth of embedding determined by the semantic structure of the compound, for example $((desk_1 \; top_3)(pen_2 \; stand_3))$, where increasing numbers stand for decreasing stress value. The CSR is essentially a reversal of the Nuclear Stress Rule (NSR) mentioned in connection with $\varphi_{phrase}$, and, like the NSR, is a bijective numerical encoding of hierarchical semantic structure as a default foundation for the assignment of phonetic prominence. This account of the CSR as being semantically based differs from mainstream accounts, but it holds if the linear combinatorics of $\tau_{word}$ and interpretation by $\sigma_{word}$ are separated.

The recursive CSR pattern, often with centre-embedding, represents a relatively direct relation between semantic-pragmatic interpretation in terms of hyponymy and 'new information' on the one hand, and $\varphi_{word}$ on the other. The new information provided by the determinans is associated with the strongest stress position, and the old information in the determinatum has a secondary stress position: it is not simply a question of plain syntactic combinatorics.

But the semantically motivated stress positioning can be overridden by higher ranking patterns. Overriding also occurs in lexicalised quasi-contrastive cases: *²Oxford ¹Road* in opposition to the canonical pattern *¹Oxford ²Street*. Lexicalisation of stress patterning is also found in German, with a canonical linear penultimate word stress position as in simplex words like *¹Doktor* and *Dok¹toren* instead of the regular compound stress position: *¹Stein ²ha gen* is the expected pattern for a village name and is used in national news broadcasts. However, the local pronunciation in and near the village is *²Stein ¹ha gen*. Similarly with supraregional *¹Pader²born* in the media, and regional *²Pader¹born* (Bleiching 1992). Another case of overriding by higher-ranking factors is the famous *thirTEEN* in final position, but *THIRteen MEN*.

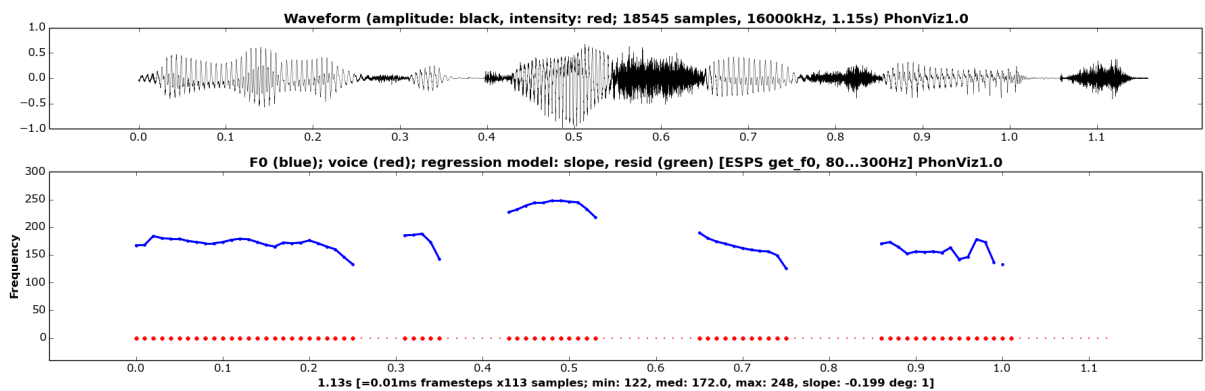

Figure 9 shows a pitch trace from an extract from the English examples of Figure 4 and Figure 7: a compound noun consisting of a derived noun and a simplex noun, *the Unification Church*. The exact phonetic form of the pitch accent associated with a word stress position in a language like English is quite variable, as is the case at the phrase rank. Two regularities are involved in describing the prosody of this particular compound:

1. The stress position is semantically predicted to be on the first word and phonetically interpreted as a pitch accent, since the word is in nuclear stress position.
2. The first word in the compound is a derivation with penultimate stress, following standard rules for stress position assignment in English words of this type, *unifi¹cation*.

The analysis of $\tau_{word}$ and $\varphi_{word}$ shows that these domains conform to the *Procedural Plausibility Condition*. Word structures in each of the subranks of morphology and phonology differ from phrasal structures, thereby also confirming the *Sui Generis Condition*. The assignment of hierarchical structure, for example by the Compound Stress Rule, is attributed to semantic relations, mainly hyponymy, rather than to grammatical constraints, and may be overridden by higher-ranking factors.

## 7.5   Prosodic-phonetic interpretation: lexical tone

Prosodic-phonetic interpretation at the word rank also applies to phonemic and morphemic lexical tones. When lexical tones occur in context, their phonetic form is conditioned by tone sandhi effects with neighbouring tones, by intonation and in some cases by constituents of the syllables associated with the tones. In many Niger-Congo tone languages, there are two contrastive lexical tones, high (H) and low (L). In many of these languages, sequences of these tones follow a regular rule: the H-L fundamental frequency movement is greater than the L-H pitch movement, a partial progressive assimilation of the pitch of the second H to that of the preceding L. In a sequence H-L-H, the second H is thus lower than the first, *automatic downstep*. The overall pattern produced by these conditions is called *tonal terracing* or *terraced tone*, because the visualisation of the overall downstepping contour has been said to resemble a descending sequence of high and low tone terraces which are relatively level in themselves. Terrace patterns must be distinguished from general declination tendencies. The term 'downstep' has also frequently been applied to describing the traditional 'stepping head' in English, but this pattern is motivated differently from terraced tonal downstep; the overall pattern is not terraced in the same sense. Terrace patterns are also often more complex than this basic case. Nevertheless, terrace relations can be described by regular grammars and finite state transducer models (Gibbon 1987, 2001, 2017).

When a low tone in an H-L-H sequence in a lexical item is deleted, the second H may retain the downstepped pitch which was formerly conditioned by the now deleted low tone, and have a lower frequency than might be expected for an H-H sequence. The missing L is frequently analysed as an abstract *floating low tone* which triggers a *lexical downstep* sequence in a sequence of two high tones: H-↓H. The context is lexical, not phonetic: the preceding lexical high tone triggers downstep, not the preceding phonetic realisation of the tone.

Downstep in Tem (Togo, ISO 639-3: *kdh*) is a little different, although there are still just two lexical tones. First, the relation affects not only downstepped high tone but also upstepped low tone; second, the assimilations of high to low and of low to high are complete, that is, a high tone is realised with approximately the pitch of the preceding low tone, and a low tone is realised with approximately the pitch of the preceding high tone. The stepping assimilation does not affect the final tone in the sequence. An example of tone sandhi in Tem, taken from Tchagbalé's tone analysis exercises (Tchagbale 1984) and his recordings of the exercises, is visualised in Figure 10.

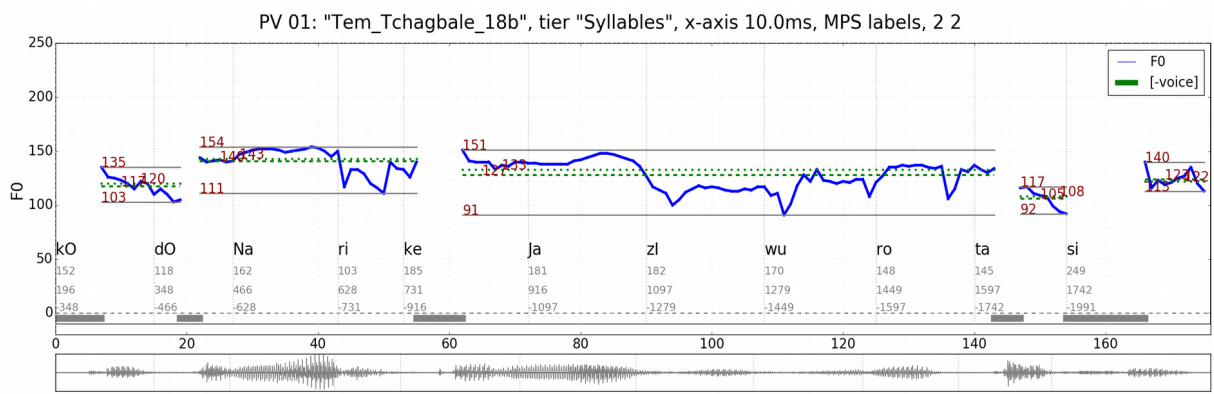

Figure 10: Tem (Niger-Congo, Gur) tone patterns in context: "kodóngariké nyazi wúro ta sí" (laughing as though the king were not dead).

Figure 10 shows the relation between syllables, lexical tone and fundamental frequency pattern for the sentence *kodóngariké nyazi wúro ta sí* (laughing as though the king had not died), where the acute accent represents a high tone and low tones are not marked. Inspection of the fundamental frequency contour, taking unevenness due to microprosody into account, shows that there is total downstep not only of lexical high tone to preceding lexical low tone, but also total upstep of lexical low tone to preceding lexical high tone, except on the final syllable.

The overall mapping is shown in Table 3. The output of the mapping function can be further interpreted as a pitch pattern, for example for use in a speech synthesiser.

Table 3: Tone sandhi mapping in Tem (↓H: downstepped high tone; ↑L: upstepped low tone; upper case: phonological tone; lower case: phonetic allotone).

| *Syllable:* | ko | dó | nga | ri | ké | nya | zi | wú | ro | ta | sí |
|---|---|---|---|---|---|---|---|---|---|---|---|
| *Lexical:* | L | H | L | L | H | L | L | H | L | L | H |
| *Mapping:* | L | ↓H | ↑L | L | ↓H | ↑L | L | ↓H | ↑L | L | H |
| *Phonetic:* | l | l | h | l | l | h | l | l | h | l | h |

The allotone mapping function shown in Table 3 can be formulated as a simple finite state transducer with three states, the initial state $State_{INIT}$, the iterating and final states $State_H$ and $State_L$, and six transitions. Upper case H and L are used for lexical tones and lower case h and l for the phonetically interpreted tones:

1.  transitions from the initial state to the high and low states:
    $<State_{INIT}, H, h, State_H>, <State_{INIT}, L, l, State_L>$
2.  terrace defining loops on each of the high and low states:
    $<State_H, H, h, State_H>, <State_L, L, l, State_L>$
3.  downstep and upstep defining transitions between the high and low states:
    $<State_H, L, h, State_L>, , <State_L, H, l, State_H>$

Table 3 and the transducer clarify that allotone conditioning derives from the preceding lexical tone, not from the preceding transduced phonetic allotone. Equivalent phonological rules can also be formulated, for example H→h/H__, L→h/H__, L→l/L__, H→l/L__.

# 8   Summary, conclusions and outlook

## 8.1   From Duality to Rank Interpretation Architecture

The main goal of the present study is to show that the core design feature of human languages is a hierarchy of finite depth consisting of ranks of signs. The signs at each rank have their own distinct ternary configuration of default linear structures, semantic-pragmatic interpretations in cognitive and cultural contexts, and externally observable prosodic-phonetic interpretations. The default property of linearity can be partially overridden within strict procedural limits in specific registers such as writing and formal rehearsed speech. To account for these properties, a *Procedural Plausibility Condition* requiring linear processing time and finite working memory was proposed. A semiotically motivated Rank Interpretation Architecture is put forward as an overall frame of reference, with six ranks, generalising from the older word rank concepts of 'duality' and 'dual articulation'. Each rank is subject to a *Sui Generis Condition* that the ranks have essentially different properties, while still being subject to the *Procedural Plausibility Condition*. An empirically realistic and procedurally plausible Multilinear Grammar approach to describing each rank was outlined.

In addition to the pervasive linearity property, the Rank Interpretation Architecture differs fundamentally from traditional frameworks for speech and language in that each rank has its own specific semiotic pair of semantic-pragmatic interpretation and prosodic-phonetic interpretation. This contrasts starkly with traditional architectures which put semantics at the 'top', syntax in the 'middle' and phonetics at the 'bottom', a conceptualisation which is often reflected in the 'back end' – 'front end' operational models of the language technologies. Different ranks may have semantic and phonetic similarities as well as sharing linear structural properties. For example, prosodic patterns such as rising terminal intonation may have similar meanings at different ranks, 'non-terminal' meaning a non-final position in an uncompleted list or a non-final subject constituent at the phrasal rank, or a question as a non-final constituent of an adjacency tuple at the discourse rank.

We are aware, at the same time, of the limitations and simplifications which are necessarily involved discussing a comprehensive framework, and of the risks involved in making essentially illustrative use of empirical examples. The framework, however plausible, will evidently remain a well-motivated heuristic methodology rather than a confirmed theory until the extensive literature is discussed in full detail and documented with more detailed observational studies at each point. There are also many aspects of the topic which have only been touched on in passing, most obviously details of semantics and pragmatics, and aspects of prosodic-phonetic interpretation which concern timing and rhythm. Some aspects of timing have been covered in the same framework by Gibbon (2017). The gestural character of phonetic interpretation (Browman and Goldstein 1989) and its relation to communicative facial and manual gesture and posture (Gibbon 2011) was not covered in any detail, nor was the application of Multilinear Grammar to the ranks of written discourse.

## 8.1 Generalisation to stochastic flat linear models

There are many potential fields of application for an integrative framework such as Multilinear Grammar and the Rank Interpretation Architecture. For example, prototypes of computational models using frameworks related to Multilinear Grammar have been developed since the 1970s (Carson-Berndsen 1998; Frazier and Fodor 1978; Kaplan and Kay 1994; Kirchhoff 1996, 1999), and future work based on these foundational results may take different directions, from descriptive and computational linguistic descriptions through stochastic grammars to deep neural networks.

Work is currently in progress on implementing the Multilinear Grammar approach in a stochastically enhanced computational model. A first step is the design of a toy application for restricted registers of English and German on the lines of the IDyOM model (Pearce 2005, Pearce et al. 2010a), which was originally developed for processing and production of music (Pearce and Wiggins 2007; Pearce et al. 2010b) of music. The implementation combines Natural Language Processing techniques with computation of information-theoretic measures and it has been demonstrated that it can handle both music and language data (Pearce and Wiggins 2007).

The algorithm builds a variable order *n*-gram model of symbolic data such as musical notes or textual representations such as phoneme strings.[7] This allows a linear model of the data to be built incrementally as it comes in, generating a sequence of probability values for each symbol encountered. Like speech prosody, music is not just a sequence of pitches, spectral shapes and phones, but a configuration of notes in sequence and in parallel, with durations, tempos, rhythms with representations at different ranks and stochastic models therefore also need to model simultaneous occurrence of perceivable events at different ranks. For language, it has been shown that segmentation performance increases when language models which include stress patterns are considered (e.g. Wiggins 2012, Griffiths et al. 2015c).

A Multilinear Grammar model can also be implemented in a viewpoint system (Pearce et al. 2005, Hedges and Wiggins 2016), which takes account of the multilinear nature of the input or output. Probabilities are calculated as symbols are consumed or emitted on one input or output stream. For the prediction of the next symbol the probabilities of several streams are taken into account and combined to calculate the probability of the next event.

## 8.2 Outlook

What, then is the benefit of the Rank Interpretation Architecture and Multilinear Grammar framework, other than as an exercise in neatness of linguistic description and in the application of Occam's Razor as a step forward from hybrid frameworks with *ad hoc* interfaces? We have presented several empirically motivated arguments for the framework. The most significant overall benefit is that the Rank Interpretation Architecture framework, together with the *Sui Generis Condition* on the ranks, provides a coherent approach to integrating components of the complex combinatorial communication system of language and speech. The framework contrasts with approaches which interface a set of hybrid components, with aporoaches which put semantics at the 'top' and phonetics at the 'bottom' with syntax in between, and with approaches which define pragmatics as a component which subsumes semantics, which in turn subsumes syntax. A subsidiary benefit is indeed the application of Occam's Razor to the hybrid collection of rule types used in linguistics. In the long term, a complete 'working model' covering all ranks would be implemented in a uniform manner, for example with neural networks.

---

7 https://code.soundsoftware.ac.uk/projects/idyom-project/wiki/Database

There are three further areas which are often treated in isolation but which potentially benefit from integrative treatment within the Multilinear Grammar and Rank Interpretation Architecture framework: prosody, the lexicon and multimodality.

The domain of prosody figures prominently in the present discussion but elsewhere is often restricted to words and sentences and treated as a quite different domain from conventional phonetic interpretation rather than being treated as an integral part of a rank architecture of prosodic-phonetic interpretation; in this respect, the Multilinear Grammar and Rank Interpretation Architecture framework is to some extent a generalisation of previous work on the prosodic hierarchy.

A second relatively isolated area concerns lexicon-like sign sets at all ranks. Sign inventories at different ranks are generally treated quite separately, from phoneme inventories through the conventional lexicon to collections of fixed expressions, idiomatic metaphors, proverbs, quotable quotes and compendia of prayers and poems for learning by heart. However, the Rank Interpretation framework provides a context for describing the inventory type at each rank, in terms both of semantic-pragmatic interpretation and of prosodic-phonetic interpretation.

A third relatively isolated area (at least from the perspective of mainstream linguistics) is the area of parallel processes in multimodal comunication at all ranks, in signing and in conversational gesture as well as in writing. These domains have properties which can be seamlessly integrated into the Rank Interpretation and Multilinear Grammar framework. There are principled links between speech and gesture which have been investigated by McNeill (2000), Kendon (2004), Gibbon (2011) and Rossini (2012), among others, and an early explicit framework for gesture within speech is provided by Articulatory Phonology (Browman and Goldstein 1989).

Above all, the Multilinear Grammar and Rank Interpretation Architecture framework poses a challenge: the integration of procedural and non-word-oriented aspects into language theory development and application in a coherent manner, and the provision of realistic evaluation criteria for language theories and models. Kindt predicted earlier (Kindt 1998:266):

> And in particular it is becoming clear that a reappraisal of the descriptive capacity of finite state systems and plausible extensions is needed. In the long run this will lead to a paradigm change in the theory of grammar and will end the Chomsky influenced epoch of the investigation of one-dimensional systems.[8]

---

8  Und speziell wird deutlich, dass eine Neubewertung der Beschreibungskapazität von Finite State-Systemen und naheliegenden Erweiterungen erforderlich ist. Sie wird über kurz oder lang zu einem Paradigmenwechsel in der Grammatiktheorie führen und die von Chomsky geprägte Epoche der Untersuchung eindimensionaler Systeme beenden.